\documentclass[10pt,journal,compsoc]{IEEEtran}

\ifCLASSOPTIONcompsoc
  \usepackage[nocompress]{cite}
\else
  \usepackage{cite}
\fi

\ifCLASSINFOpdf
   \usepackage[pdftex]{graphicx}
\else
\fi

\usepackage{amsmath,amssymb,amsfonts,amsthm}

\usepackage{url}

\usepackage{cite}
\usepackage[linesnumbered,ruled,vlined]{algorithm2e}
\usepackage{graphicx}
\usepackage{multicol,multirow}
\usepackage[usenames,dvipsnames]{xcolor}
\usepackage{textcomp}
\usepackage{adjustbox}
\usepackage{diagbox}
\usepackage{booktabs}
\usepackage{lettrine}
\usepackage{siunitx}
\usepackage{caption,subcaption}
\usepackage{tcolorbox}
\usepackage{enumitem}
\usepackage{soul}
\usepackage{cancel}
\usepackage{balance}
\usepackage{flushend}

\let\oldproofname=\proofname
\renewcommand{\proofname}{\rm\bf{\oldproofname}}
\newlength{\twosubht}
\newsavebox{\twosubbox}
\begin{document}
\title{Practical Mission Planning for Optimized UAV-Sensor Wireless Recharging}

\author{Qiuchen~Qian, James~O’Keeffe, Yanran~Wang~and~David~Boyle%
}

\IEEEtitleabstractindextext{%
\begin{abstract}
Optimal maintenance of sensor nodes in a Wireless Rechargeable Sensor Network (WRSN) requires effective scheduling of power delivery vehicles by solving the Charging Scheduling Problem (CSP). Deploying Unmanned Aerial Vehicles (UAVs) as mobile chargers has emerged as a promising solution due to their mobility and flexibility. The CSP can be formulated as a Mixed-Integer Non-Linear Programming problem whose optimization objective is maximizing the recharged energy of sensor nodes within the UAV battery constraint. While many studies have demonstrated satisfactory performance of heuristic algorithms in addressing specific routing problems, few studies explore online updating (i.e., mission re-planning `on the fly') in the CSP context. Here we present a new offline and online mission planner leveraging a first-principles power consumption model that uses real-time state information and environmental information. The planner, namely \textbf{R}apid \textbf{O}nline \textbf{M}etaheuristic-based \textbf{P}lanner (ROMP), supplements solutions from a Guided Local Search (GLS) with our Context-aware Black Hole Algorithm. Our results demonstrate that ROMP outperforms GLS in most cases tested. We developed and proposed FastROMP to speed up the online mission (re-)planning algorithm by introducing a new online adjustment operator that uses the latest state information as input, eliminating the need for re-initialization. FastROMP not only provides a better quality route, but it also significantly reduces computational time. The reduction ranges from 39.57\% in sparse deployment to 93.3\% in denser deployments.
\end{abstract}

}

\maketitle

\IEEEdisplaynontitleabstractindextext

\IEEEpeerreviewmaketitle

\IEEEraisesectionheading{\section{Introduction}\label{sec:introduction}}
\IEEEPARstart{W}{}ireless Rechargeable Sensor Network (WRSN) systems are an emerging paradigm within Internet-of-Things (IoT) technologies~\cite{FAN2018WRSNSURVEY}. Maintaining a field of wireless sensors can be challenging due to the need for human intervention and costly infrastructure support. One solution to this challenge is deploying autonomous vehicles to perform maintenance tasks, such as mobile remote charging~\cite{LIN20213DCHAGRING}. However, efficiently servicing large-scale networks using autonomous vehicles is a complex navigational problem. This paper addresses mobile chargers' dynamic Charging Scheduling Problem (CSP) in WRSNs. Autonomous Unmanned Aerial Vehicles (UAVs), such as multi-rotor copters, are well-suited to this task in large-scale WRSN scenarios due to their agility and flexibility~\cite{MITCHEN2017WPT}. UAVs can use Inductive Power Transfer (IPT) to recharge sensor nodes rapidly~\cite{Boyle_IC_2016, MITCHEN2018WPT}. We approach the CSP as a multi-objective optimization problem where the UAV must wirelessly recharge a specific number of sensor nodes. The goal is to maximize the recharged energy of all devices deployed across a field while minimizing the UAV's discharged energy. We define two sub-problems from different perspectives to achieve the overall optimization objective: 
\begin{itemize}
    \item Travelling Salesman Problem (TSP): A single UAV is assigned to charge all unvisited sensors in a given WRSN. Multiple iterations may be required due to the UAV battery's limited capacity. The optimization objective of a single iteration is to minimize the UAV's discharged energy.
    \item Orienteering Problem (OP): A single UAV is assigned to charge all unvisited sensors in a given WRSN. Each sensor is associated with a biased `prize' weight. This model allows a temporary drop of some low-prize sensors. The optimization objective is to maximize the collected prizes, subject to a budget constraint on UAV's discharged energy.
\end{itemize}

\begin{figure}[!t]
    \centering
    \begin{subfigure}[htbp]{0.241\textwidth}
        \includegraphics[width=\textwidth]{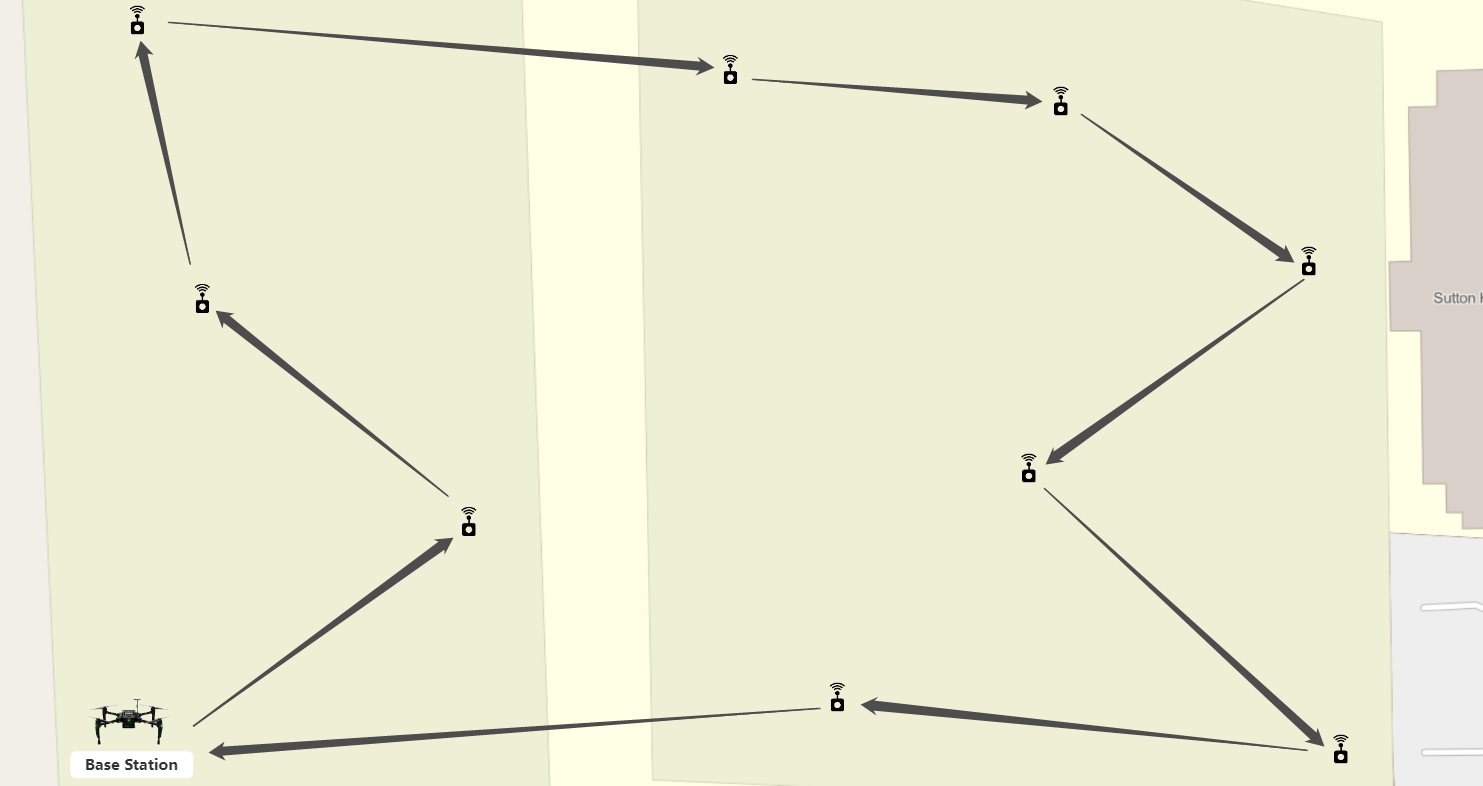}
        \label{fig:graph_intro_tsp}
    \end{subfigure}
    \begin{subfigure}[htbp]{0.241\textwidth}
        \includegraphics[width=\textwidth]{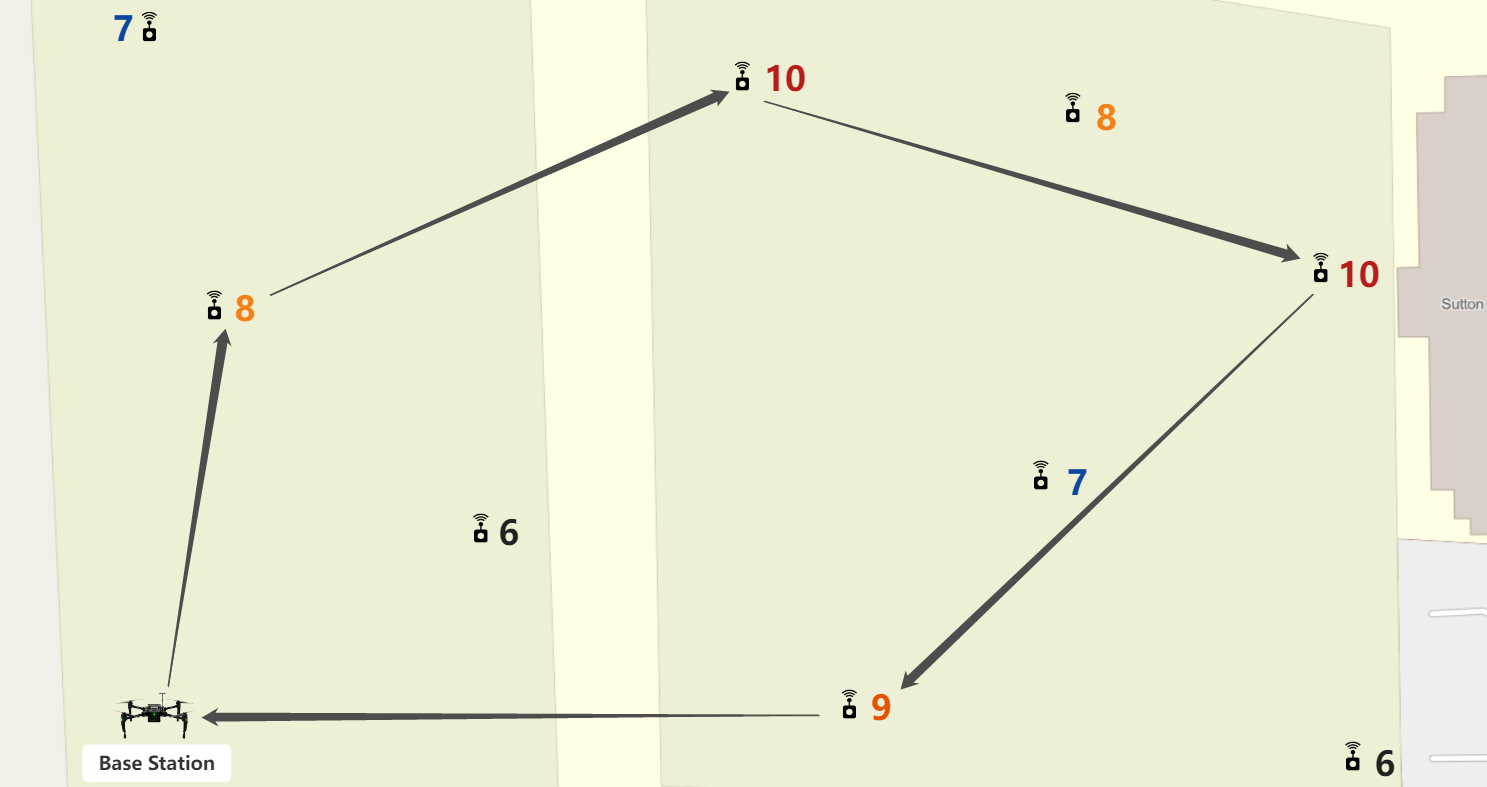}
        \label{fig:graph_intro_op}
    \end{subfigure}
    \caption{Graphical representation of TSP (left) and OP (right) in a UAV Charging Scheduling Problem.}
    \label{fig:graph_intro}
\end{figure}

The TSP and OP (see Fig.~\ref{fig:graph_intro}) are classified as NP-hard problems, which means finding an optimal solution requires an exhaustive search of all possible solutions. While exact and approximation algorithms can guarantee the solution's optimality~\cite{fischetti1998solving}, their practical computational expense increases significantly with the increase of the instance scale. Therefore, heuristic algorithms that can quickly converge to (near-)optimal solutions are generally preferred in most computation-sensitive scenarios as they empirically show a good balance of computation time and solution quality~\cite{VES2020ML4COREVIEW}. Single-solution-based meta-heuristics, such as Simulated Annealing (SA), Tabu Search (TS) and Guided Local Search (GLS)~\cite{HAH2019HEURREVIEW}, modify and improve single elements in the candidate solution. In contrast, population-based meta-heuristics, such as Genetic Algorithm (GA), Ant Colony Optimization (ACO) and Black Hole Algorithm ~\cite{HAH2019HEURREVIEW, HAT2018BHA}, search for potential solutions by modifying multiple elements simultaneously.

A hybrid offline and online scheme is typical for optimizing UAV path planning in a dynamic environment, e.g., obstacle avoidance~\cite{YIN2018OFFLINEONLINEUAV, WANG2022KINOJGM}. The offline algorithm searches for a path that shortens travel time and avoids obstacles from a static safety index map. The online algorithm quickly adjusts the offline path to avoid dynamic or unexpected obstacles from a dynamic map. Similarly, due to the dynamic nature of the environment in which UAVs operate and the presence of unforeseen adversarial factors such as turbulence or strong winds, it is impossible to solve the CSP precisely, using only static information. Therefore, an online approach that can adapt to changing environmental conditions is necessary to complete the UAV's mission successfully. 

Moreover, an appropriate power consumption model for the UAV is crucial in evaluating its capability to complete missions under varied scenarios. Without prior knowledge of the UAV’s capabilities, the power model can effectively translate the UAV's actions into actual energy consumption. This helps the algorithm to better formulate the fitness function. Efforts to analyze power consumption models of multi-rotor UAVs operating in dynamic environments are regularly reported, e.g., in~\cite{YC2018UAVENERGY, LIU2017UAVPOWER, THI2021M100DELIVERY}. Many studies use anemometers to measure wind field variations, which are commonly recognized as the primary 
influencing factor on the power consumption model in such environments. To record real-time wind rose and relative instantaneous power, Rodrigues~\emph{et al.} deployed an anemometer and a power sensor on the DJI Matrice 100 (M100) drone~\cite{THI2021M100DELIVERY}. Their work indicates extending the planner from offline to online planning is possible by applying dynamic updates, e.g., time-variable wind data and remaining battery capacity. Mission planning should generally occur during the time interval of interaction between the UAV and the sensor node to avoid additional strain on onboard computing resources during real-time path and trajectory planning. The interaction time depends on the time required to charge the energy container, such as a battery or supercapacitor. The IPT transmitter described by~\cite{ARTEAGA2019WPT, ALDHAHER2017LIGHTWEIGHTWPT} is capable of powering loads of up to 150 W. Therefore, a small supercapacitor (e.g. 3 Farads) can be rapidly charged within seconds, while charging a Li-ion battery can take tens of minutes. 

Our key contributions can be summarized as follows: 
\begin{itemize}
\item We study autonomous UAV-assisted energy supply management of a WRSN in dynamic environments. We provide a Mixed-Integer Non-Linear Programming formulation for the CSP. To tackle the CSP, we develop an offline and online mission planning scheme and a first-principle power consumption model of different UAV regimes in a windy environment: takeoff, landing, cruise and charging. 
\item We present ROMP, a novel approach that combines GLS with a Context-aware BHA to generate an initial charging route for the UAV operating in dynamic environments. To our knowledge, this is the first paper to extend classic BHA in addressing the CSP. Our results demonstrate that ROMP can effectively leverage wind conditions to enhance GLS solution performance even in small-scale instances where GLS is highly likely to yield an optimal route. Our results (see \S~\ref{sec:iter}) demonstrated further improvements in ROMP.
\item We demonstrate the feasibility of deploying ROMP online, however, the computational time often remains unsatisfactory in practice. We therefore developed and present FastROMP, a novel computationally efficient mission (re-)planning approach. FastROMP incorporates dynamic drop and add operators with CBHA to readjust the previous plan based on the latest parameter updates, omitting the need for re-initialization. Our results (see \S~\ref{sec:dync}) indicate that FastROMP can maximize energy budget utilization and charge more energy with significantly reduced computational time compared to ROMP. 
\item We present a detailed performance analysis of ROMP and FastROMP under various scenarios. We also examine the parallel search efficiency of CBHA, which provides a general guide for practitioners to apply this planner in \textit{ad hoc} scenarios.
\item Software implementation and experimental results of ROMP are made openly available to the community\footnote{[Online] \url{https://github.com/sysal-bruce-publication/ROMP.git}}.
\end{itemize}

\section{Related Work}
\label{sec:related-work}
Recent advancements in Wireless Power Transfer technology have led to an increased interest in WRSNs. The CSP has emerged as a significant area of research within the field of mission planning for WRSNs. Various studies have employed different modelling strategies and optimization techniques to address the CSP. A review of the relevant literature is presented below.

\subsection{Different Modelling Strategies to Address the CSP}
Lin~\emph{et al.} modelled a three-dimensional CSP as a TSP with optimization objectives of determining the best charging locations and charging duration~\cite{LIN20213DCHAGRING}. While the TSP model simplifies system complexity and generally converges on relatively low-cost solutions, its uniform charging priority could lead to sensor data loss in practice, particularly in long-term monitoring. Pandiyan~\emph{et al.} addressed a two-stage wireless recharging problem with a weighted clustering GA and the TSP model during the solution search phase~\cite{PAN2020ODRS}. Qian~\emph{et al.} extended this approach by assigning tasks to multiple UAVs, using a K-Means Clustering initialization. This allows a flexible number of vehicle-node visitations in each route~\cite{QQ2021ORS}. Simulated results showed that their proposed BHA delivered a comparable performance with faster execution than a state-of-the-art GA. However, these clustering strategies did not account for the charging capability of mobile agents, potentially resulting in infeasible missions being assigned to energy-constrained vehicles. Lin~\emph{et al.} proposed an on-demand charging architecture to involve non-deterministic factors in the CSP~\cite{LIN2018TSCA}. Their node deletion and insertion algorithm allowed for flexible adjustment of the current charging queue by avoiding low-efficient nodes and saving dying nodes - similar to the advantages of the OP model.

\subsection{Practical Difficulties in UAV Mission Planning}
Generating an efficient and safe schedule for a UAV-assisted WRSN is a challenging task. Li~\emph{et al.} addressed this issue by incorporating the UAV's energy model into their full and partial data collection maximization problem. However, they assumed a fixed hovering and travelling energy consumption rate in the energy model~\cite{SOH2019FIXEDCONSUMPTIONRATE}. Thibbotuwawa~\emph{et al.} incorporated weather forecast data into the delivery planning of a UAV fleet, which assumed globally constant wind within a time window~\cite{THI2019UAVDELIVERY}. Though the solution in~\cite{THI2019UAVDELIVERY} could update online wind speed periodically, it is unreliable because the UAV could not react to a stochastic event such as unforeseen energy deficiency. Dorling~\emph{et al.} employed SA with a simplified energy consumption model for the multi-rotors to address a dynamic drone delivery problem~\cite{Dorling2017VRPDELIVERY}. Their experimental results highlighted the significance of an appropriate energy model in UAV mission planning. Shi~\emph{et al.} proposed a two-stage approach that combined a UAV Motion Control (UMC) algorithm with a Dynamic Genetic Clustering (DGC) algorithm to search for target sensors and recharge them in unknown environments~\cite{SHI2023TWOSTAGE}. The UMC and DGC algorithms were designed to maximize the number of sensors discovered by the UAV without prior location information and to optimize the recharged energy for the WRSN. However, unforeseen adversarial factors during the `Power Transfer' phase were not explicitly considered, potentially resulting in the charging plan generated by the DGC algorithm being infeasible and unnecessary energy waste during the `Target Search' phase. At the time of writing, we are unaware of any other study focused on solving our specific CSP for a single UAV that incorporates dynamic environmental effects into solution performance and mission safety~\cite{HUANG2018DRONECHARGINGSURVEY, THI2018ENERGYREVIEW, KHO2019TSPPVRPUAVSURVEY}. Some previous research applies approximation algorithms to solve their particular problems. These algorithms can provide a theoretical guarantee by ensuring the solution quality is better than a lower bound. However, the time complexity of approximation algorithms for tackling the dynamic CSP is generally too high for our purposes~\cite{XU2020APPROXIMATIONALGS, LI2021DATACOLLECTION, LIN20213DCHAGRING}. We aim to efficiently compute an initial route and optimize its quality for nonlinear system models with online updating capability. Therefore, we apply meta-heuristic algorithms to yield feasible solutions with reasonable latency.

\section{System Power Consumption Model}
\begin{figure}[!b]
    \hspace*{-1.5mm}  
    \includegraphics[width=0.5\textwidth]{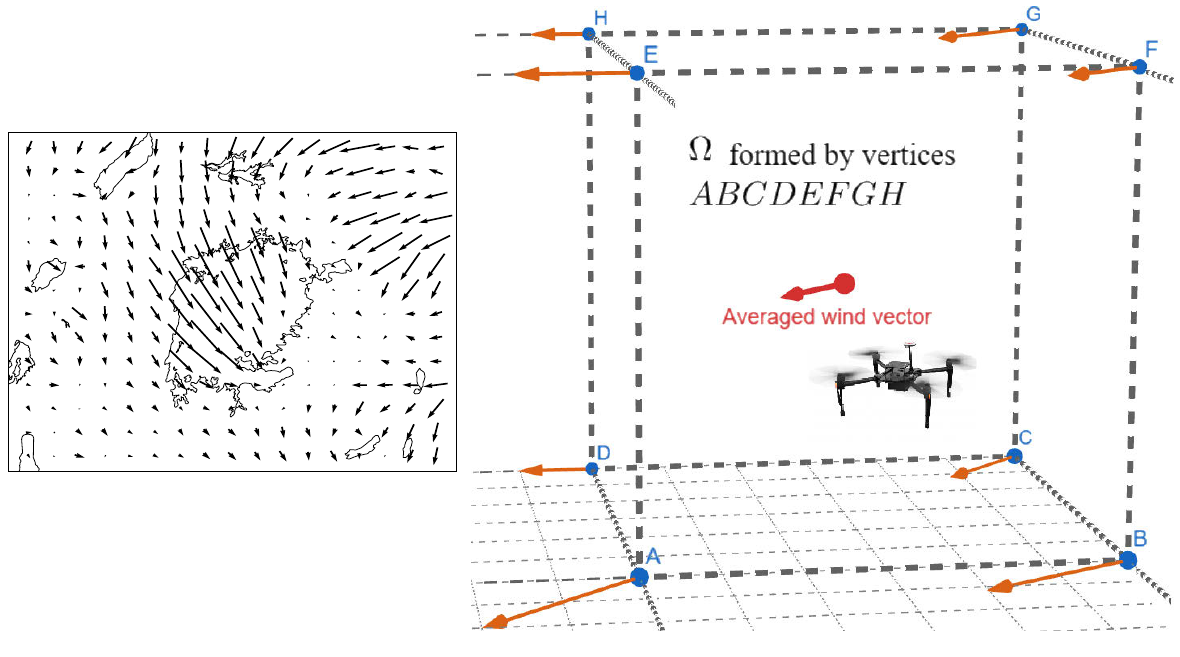}
    \caption{An example 2D quiver plot of wind speed over a lake, where the UAV traverses in a wind cube region $\Omega$, formed by eight vertices $A, B, C, D, E, F, G$ and $H$. The wind velocity vector of $\Omega$ is averaged over eight wind velocity vectors at these vertices.}
    \label{fig:wind-cube}
\end{figure}
The battery capacity of a UAV restricts its ability to recharge. An appropriate model for UAV energy consumption is essential to evaluate the quantity and distribution of target sensor nodes that need to be covered. In this section, we derive equations for the power consumed by a quadrotor drone during the mission. 

The UAV energy consumption can be significantly affected by wind. Integrating wind and UAV aerodynamic models can be computationally expensive, and a high-precision estimate of discharged energy is unnecessary for our planner to obtain a near-optimal solution. Therefore, we employed a model of reduced order complexity. To balance the accuracy and computational complexity of our proposed model, assumptions are made following~\cite{liu2017energy}:
\begin{enumerate}
    \item We discretize the wind field into a time-variant 3D lattice, where each vertex $p_i$ has a wind velocity vector $V_{wind}(t, p_i)$. Eight adjacent vertices form a wind cube region $\Omega$, whose edge length determines the resolution of the wind field (see Fig.~\ref{fig:wind-cube}). Let $\mathbf{x}(t)$ be the coordinate of the UAV at time $t$. We assume the UAV is affected by a constant wind $\mathbf{V}_{wind}(\mathbf{x}(t)) = \sum_{i=1}^8 \mathbf{V}_{wind}(t, p_i) / 8$, \textbf{iff} $\mathbf{x}(t)$ is in the same region $\Omega$. Therefore, the UAV flight regimes can be approximated to steady-state motion, which mainly considers thrust force, gravity force, and frictional drag force of surrounding airflow~\cite{YC2018UAVENERGY}. 
    \item The computation will be performed offline in advance, and thus the wind field forecast is reasonably accurate. Because the mission duration is relatively short (typically less than 30 minutes~\cite{M100, liu2017energy}), the magnitude error of actual energy consumption caused by offline and up-to-date wind fields is trivial, and it can be neglected. As the mission progresses, this assumption remains valid on the accumulated error of sensor nodes' rechargeable energy. 
    \item We categorize UAV actions during the mission into four distinct regimes: takeoff, landing, cruise, and charging, where takeoff and landing are about vertical movement; cruise is about horizontal movement; The UAV remains static in charging. The power consumption is mainly constituted by induced power $P_i$, profile power $P_p$, parasite power $P_{par}$ and ancillary power $P_{a}$ for flight regimes~\cite{RODRIGUES2022100569}. To reduce the non-dominant computation in a dynamic environment (i.e., wind field), we perform a first-principles analysis to estimate $P_i$ and $P_{par}$ only. To estimate the time required for the UAV to reach a target location, we pre-specify a constant UAV ground speed $V_{ground}$, takeoff speed and landing speed. The UAV traverses from the current node to the next node, following takeoff $\to$ cruise $\to$ landing. 
    \item For the majority of the mission, the UAV is in a steady state, whereby force equilibrium is maintained. Therefore, the vertical thrust is equal to the sum of gravity and sheer drag force $F_{T, vert} = F_G + F_{D, vert}$; the horizontal thrust is equal to horizontal drag force $F_{T,horz} = F_{D,horz}$.
    \item The energy drop rate of each sensor type is assumed to be linear, given that its sampling rate remains fixed. The service time of each deployed sensor node in the WRSN is estimated every time the UAV completes a mission to create a new charging list. The UAV periodically charges target nodes in the designated charging list. Considering a real-world scenario, a UAV will attempt to charge every node scheduled. This information will be reported if the attempt fails, and a manual inspection should be requested. 
\end{enumerate}

\begin{table}[!t]
    \caption{Used Variable Nomenclature in UAV Dynamics}
    \label{tab:UAV-DYNAMICS}
    \begin{adjustbox}{width=0.486\textwidth}
    \begin{tabular}{c c c}
    \toprule
    \textbf{Variable} & \textbf{Definition} & \textbf{Units} \\
    \hline
    $F_T$ & Thrust provided by motor rotation & [N] \\
    $F_D$ & Drag force experienced by the PDV & [N] \\
    $F_G$ & Gravity force experienced by the PDV & [N] \\
    $\rho_a$ & Air density & [kg/m$^3$] \\
    $c_D$ & Drag coefficient & / \\
    $A_{a}$ & Reference area of airflow & [m$^2$] \\
    $v_{a}$ & Airflow velocity relative to the propeller & [m/s] \\
    $P_{m}$ & Mechanical power of motor rotation & [W] \\ 
    $m$ & Total PDV mass & [kg] \\
    $g$ & Gravity acceleration & [m/s$^2$] \\
    \bottomrule
    \end{tabular}
    \end{adjustbox}
\end{table}

\begin{figure}[!b]
    \centering
    \includegraphics[width=0.486\textwidth]{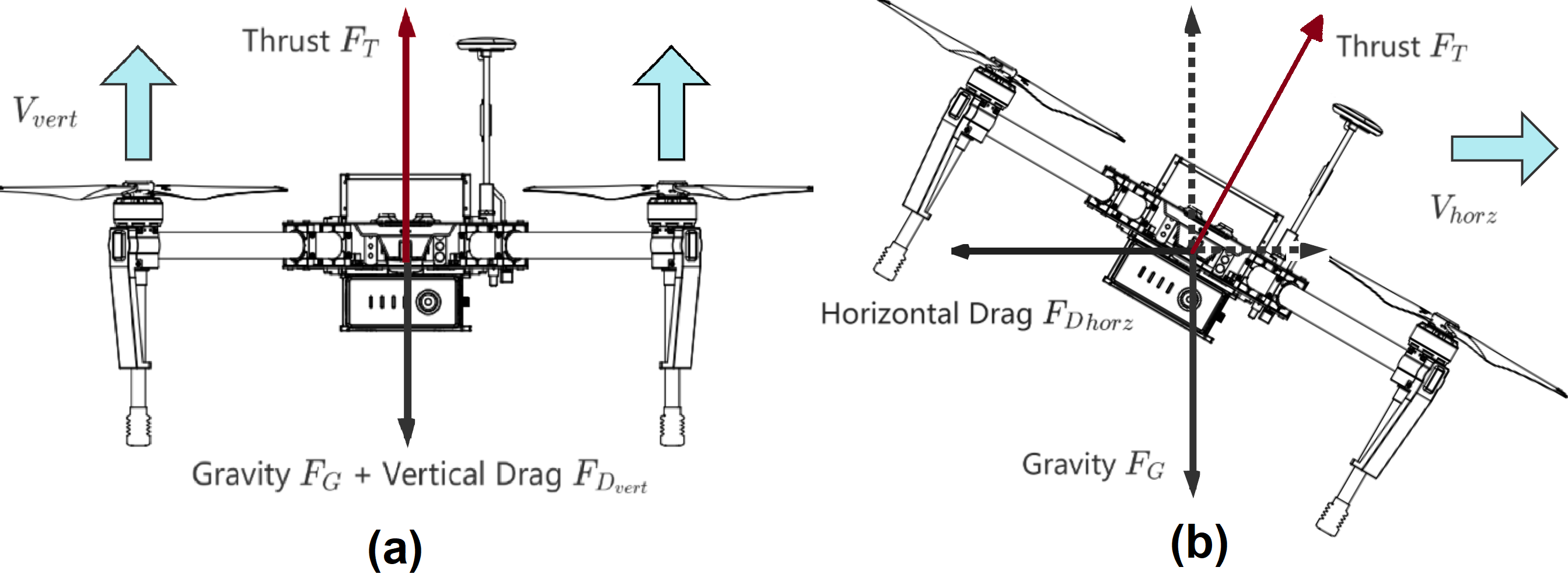}
    \caption{Vertical motion (a) and horizontal motion (b) of the M100 drone~\cite{M100}.}
    \label{fig:m100-move}
\end{figure}

First, we develop a theoretical power consumption model of UAV flight regimes. The list of modelling variables is given in Table.~\ref{tab:UAV-DYNAMICS}. The gravity force is $F_G = m g$, where $m$ is the UAV total mass, and $g$ is the gravity acceleration. Induced power $P_i$ can be derived from hover in a no-wind condition~\cite{RODRIGUES2022100569}:
\begin{equation}
\label{eq:induced-power}
    P_i = F_T V_i = \sqrt{\frac{F_T^3}{2\:\rho_a\: A_{prop}}}
\end{equation}
where $F_T$ is the thrust (here it equals to $F_G$), $V_i$ is the induced velocity, $A_{prop}$ is the area swept by all propellers. Then we introduce wind into such a system. The drag force experienced by the UAV due to a movement through constant wind can be modelled as:
\begin{equation}
\label{eq:drag-force}
    F_D = \frac{1}{2}\:\rho_a\:c_D\:A_{ref}\:V_{air}^2 
\end{equation}
where $\rho$ is the air density, $C_D$ is the drag coefficient, $A_{ref}$ is the reference area on the UAV body frame, $V_{air}$ is the UAV air velocity. Therefore, we can derive $P_{par}$ as:
\begin{equation}
\label{eq:parasite-power}
    P_{par} = F_D V_{air} = \frac{1}{2}\:\rho_a\:c_D\:A_{ref}\:V_{air}^3
\end{equation}
In a steady state, therefore, the approximate power of different flight regimes required to induce thrust can be achieved through force equilibrium, which is shown in Fig.~\ref{fig:m100-move}:
\begin{equation}
    \label{eq:all-power}
    P_f = \begin{cases}
        (F_{D, vert} + mg)^{\frac{3}{2}} / \sqrt{2 \rho_a A}\:, & \text{takeoff}\\
        (mg - F_{D, vert})^{\frac{3}{2}} / \sqrt{2 \rho_a A}\:, & \text{landing}\\
        (F_{D, horz}^2 + (mg)^2)^{\frac{3}{4}} / \sqrt{2 \rho_a A}\:, & \text{cruise}
    \end{cases}
\end{equation}

Moreover, the UAV energy consumption caused by an IPT process $E_{IPT}$ is significant. IPT is performed between the coils installed on the UAV and the sensor node. The IPT link efficiency $\eta_{IPT}$ for each sensor node may vary depending on factors such as the coil diameter, the distance between coils, and disturbances in the air. Although fluctuations in these parameters can influence overall power consumption and are relevant to our work, they are beyond the scope of this paper. Thus, we set a uniform transfer efficiency to simplify computation. $E_{IPT}$ can be derived from the energy requirement of a target sensor node:
\begin{equation}
    \label{eq:ipt}
    E_{IPT}(v_i) = \frac{C\:(V_{max}(v_i) - V_{now}(v_i))^2}{2 \eta_{IPT}} 
\end{equation}
where $C$ denotes the capacitance, $V_{max}$ and $V_{now}$ are the target sensor's maximum and the present voltage levels.

\section{Problem Formulation}
\label{sect:problem-definition}
This paper addresses a dynamic CSP whereby a single UAV is assigned to maintain the power supply of a WRSN in a dynamic environment. The priority for the UAV to recharge a sensor node depends on the sensor node's remaining energy, whereas a sensor node with lower remaining energy has a higher priority. Because we assume a sensor node's rechargeable energy is fixed, we use an integer `prize' to normalize all magnitudes. The problem can be defined on a directed complete graph $\mathcal{G} = \langle \mathcal{V}, \mathcal{D}, \mathcal{P}\rangle$, where $\mathcal{V} = \{v_1, ..., v_i, ..., v_N\}$ denotes the set of mission start node $v_1$, mission end node $v_N$ and sensor nodes $v_{i}$; $\mathcal{D} = \{ d(v_i, v_j), ... \}$ refers to the distance set, where $d(v_i, v_j)$ is the traverse distance from node $v_i$ to node $v_j$, $i \neq j$; $\mathcal{P} = \{ p(v_1), ..., p(v_i), ..., p(v_N) \}$ is the set of sensor nodes' prizes can be collected, where typically $p(v_1) = p(v_N) = 0$. Note that we assume each sensor node with a fixed prize during a single mission because the actual magnitude of energy drop is relatively small. For example, the temperature sensor LMT84 has a typical 3.3 $V$ supply with 5.4 $\mu A$ power consumption and power-on time of 0.7 $ms$~\cite{LMT84}, which means the energy dissipation will be less than 1 $\mu J$ for a 30-minute mission if we set the sampling rate a once per minute. Let $\overline{v_i v_j}$ be the path that the UAV traverses from $v_i$ to $v_j$; we then denote the set of wind cube regions covered by a route $\overline{v_i v_j}$ as $\mathcal{O}(t, \overline{v_i v_j}) = \{\Omega_1(t), ... | \overline{v_i v_j}\}$, where $|\mathcal{O}(t, \overline{v_i v_j})| \geq 1$. Because the CSP has decision variable $X_{ij} \in \{0, 1\}$, continuous variable $t$, and nonlinear constraints, we formulate the CSP to a Mixed-Integer Non-Linear Programming (MINLP) problem as below:

\begin{align}
\label{eq:obj}
&\max ~ \sum_{i=1}^{N - 2} \sum_{j=2}^{N - 1} p(v_j) \cdot X_{ij} \\
\label{con:start-end}
\textbf{s.t.}& ~\sum_{j=2}^{N} X_{1j} = \sum_{i=1}^{N - 1} X_{iN} = 1 \\
\label{con:only-1-connnect}
&\sum_{i=1}^{N-1} X_{ik} = \sum_{j=2}^{N} X_{kj} \leq 1,~~ k = 2, ..., N - 1\\
\label{con:subtour1}
&u_i - u_j + 1 \leq (N - 1)(1 - X_{ij}),~~ i, j = 2, ..., N \\ 
    \label{con:budget}
    &\begin{aligned}
    \sum_{i=1}^{N - 1} &\sum_{j=2}^{N} \biggl( \int_{t_{v_i}}^{t_{v_j}} P_f \Bigl( \mathbf{V}_{air}(\mathbf{x}(t)) \Bigr)\: dt \\
    &+ E_{IPT}(v_i) \biggr) \cdot X_{ij} \leq \mathcal{B},~~ \mathbf{x}(t) \in \mathcal{O}(t, \overline{v_i v_j})
    \end{aligned}\\
\label{con:air-speed}
& \mathbf{V}_{air}(\mathbf{x}(t)) = \mathbf{V}_{ground} - \mathbf{V}_{wind}(\mathbf{x}(t)),~~ \mathbf{x}(t) \in \Omega\\
\label{con:visit-time}
&||t_{v_j} - t_{v_i}|| \geq d(\overline{v_i v_j}) \cdot \mathbf{V}_{ground}^{-1} \\
\label{con:prize}
& lb \leq p(v_i) \leq ub,~~ lb, ub \in \mathbb{Z}^+ \\
\label{con:subtour2}
&2 \leq u_i \leq |\mathcal{V}|,~~ i = 2, ..., N \\
\label{con:safety}
& 0 < \mathcal{B} \leq \mu \cdot E_{bat}
\end{align}
where the objective function \eqref{eq:obj} maximizes the total collected prize, where the prize has an integer lower bound $lb$ and upper bound $ub$ to scale the magnitude difference (see constraint \eqref{con:prize}). Constraint \eqref{con:start-end} ensures the solution route starts at node $v_1$ and ends at node $v_N$. Constraint \eqref{con:only-1-connnect} ensures the connectivity of the solution route, and one node is visited at most once. Constraint \eqref{con:subtour1} and \eqref{con:subtour2} prevent subtours, according to the Miller-Tucker-Zemin formulation of the TSP \cite{miller1960integer}. Constraint \eqref{con:budget} limits the total energy consumption is less than the energy budget $\mathcal{B}$. Note that $E_{IPT}(v_i)$ can be calculated from Eq. \eqref{eq:ipt}, and $P_f$ can be obtained from Eq. \eqref{eq:all-power} depending on different flight regimes. Constraint \eqref{con:air-speed} enforces the UAV air velocity to be static (i.e., steady state) \textbf{iff} the UAV traverses in the same wind cube region $\Omega$. Constraint \eqref{con:visit-time} keeps track of the elapsed time from the UAV leaves node $v_i$ to the UAV leaves node $v_j$. Constraint \eqref{con:safety} is set for safety reasons (the factor $\mu$ is typically set to 80\%), which ensures the UAV has enough energy to execute the Return-To-Home process.

Note that the UAV can charge all sensor nodes in some scenarios (e.g., windless weather or a short charging list). Therefore, $v_1 = v_N \Rightarrow \mathcal{V} = \mathcal{V} \setminus \{ v_N \}$. The above MINLP formulation can be simplified as below:
\begin{align}
\label{eq:obj-tsp}
&\min ~ \sum_{i=1}^{N - 1} \sum_{j=1}^{N - 1} \biggl( \int_{t_{v_i}}^{t_{v_j}} P_f \Bigl( \mathbf{V}_{air}(\mathbf{x}(t)) \Bigr) dt \biggr) \cdot X_{ij}\\
\label{con:tsp-same-start-end2}
\text{s.t.} & \sum_{i=2}^{N-1} X_{i1} = \sum_{j=2}^{N-1} X_{1j} = 1\\
\label{con:tsp-one-visit}
&\sum_{i=1}^{N-1} X_{ik} = \sum_{j=1}^{N-1} X_{kj} \leq 1,~~ k = 2, ..., N - 1\\
\label{con:tsp-subtour}
&u_i - u_j + 1 \leq (N - 2)(1 - X_{ij}),~~ i, j = 2, ..., N-1 \\ 
\label{con:tsp-subtour2}
&2 \leq u_i \leq |\mathcal{V}|,~~ i = 2, ..., N - 1 \\
\nonumber
& \text{Constraints \eqref{con:air-speed}, \eqref{con:visit-time}}
\end{align}
where the objective function~\eqref{eq:obj-tsp} minimizes the energy consumption caused by flight regimes only. Constraint~\eqref{con:tsp-same-start-end2} represents that the route starts and ends at the same node. Constraint~\eqref{con:tsp-one-visit} enforces all sensor nodes are visited exactly once. Constraint \eqref{con:tsp-subtour} and \eqref{con:tsp-subtour2} prevent subtours.

\section{System Design}
To simulate a general mission planning scenario of UAV-assisted WRSN, we select temperature sensor LMT84~\cite{LMT84} with capacitor HB1030-2R510~\cite{HB1030} and pressure sensor NPA300~\cite{NPA300} with capacitor PHB-5R0~\cite{PHB5R0} as the sensor prototypes, following~\cite{PAN2020ODRS, QQ2021ORS}. Specifications of sensor nodes and the UAV are listed in Table.~\ref{tab:node}. Values of some parameters in Eq.~\eqref{eq:induced-power}-\eqref{eq:ipt} that experiments can identify are summarized as below: air density $\rho = 1.25kg/m^3$, drag coefficient $c_D = 0.04$~\cite{RODRIGUES2022100569}, and IPT link efficiency $\eta_{IPT} = 50$\%~\cite{PAN2020ODRS}. Although fluctuations in these parameters can influence overall power consumption and are thus relevant to our work, they are beyond the scope of this paper. Therefore, we set constant values to these parameters.

\begin{table}[!t]
\centering
\Large
\caption{Specifications of the sensor node unit and the UAV}
\label{tab:node}
\begin{adjustbox}{width=0.486\textwidth}
\begin{tabular}{cccc}
\toprule
& \textbf{Parameter} & \textbf{Temperature Sensor} & \textbf{Pressure Sensor} \\
\hline
\multirow{5}{*}{Sensor} & Model & LMT84 \cite{LMT84} & NPA-300 \cite{NPA300} \\
\cline{2-4}
& Supply Voltage (Min.) & {1.5 V} & 3.3 V \\
\cline{2-4}
& Supply Voltage (Max.) & {5.5 V} & 5 V \\
\cline{2-4}
& Typical Current & {5.4 $\mu$A} & {1.2 mA} \\
\midrule
\multirow{3}{*}{Capacitor} & Model & HB1030-2R5106 \cite{HB1030} & {PHB-5R0} \cite{PHB5R0} \\
\cline{2-4}
& Capacitance & 6 F & 3 F \\
\cline{2-4}
& {Voltage} & 2.5 V & 5 V \\
\midrule
\multirow{7}{*}{UAV \cite{M100}} & {Maximum speed of ascent} & \multicolumn{2}{c}{5 m/s} \\
\cline{2-4}
& {Maximum speed of descent} & \multicolumn{2}{c}{4 m/s} \\
\cline{2-4}
& {Maximum speed (GPS mode)} & \multicolumn{2}{c}{17 m/s} \\
\cline{2-4}
& {Full battery energy (one TB47D battery)} & \multicolumn{2}{c}{99.9 Wh} \\
\cline{2-4}
& {Windward area (horizontal and vertical)} & \multicolumn{2}{c}{0.153 $m^2$ and 0.779 $m^2$} \\
\cline{2-4}
& {Mass (include body and payload)} & \multicolumn{2}{c}{3.107 kg} \\
\bottomrule
\end{tabular}
\end{adjustbox}
\end{table}

\subsection{System Architecture}
\begin{figure}[!b]
\includegraphics[width=0.486\textwidth]{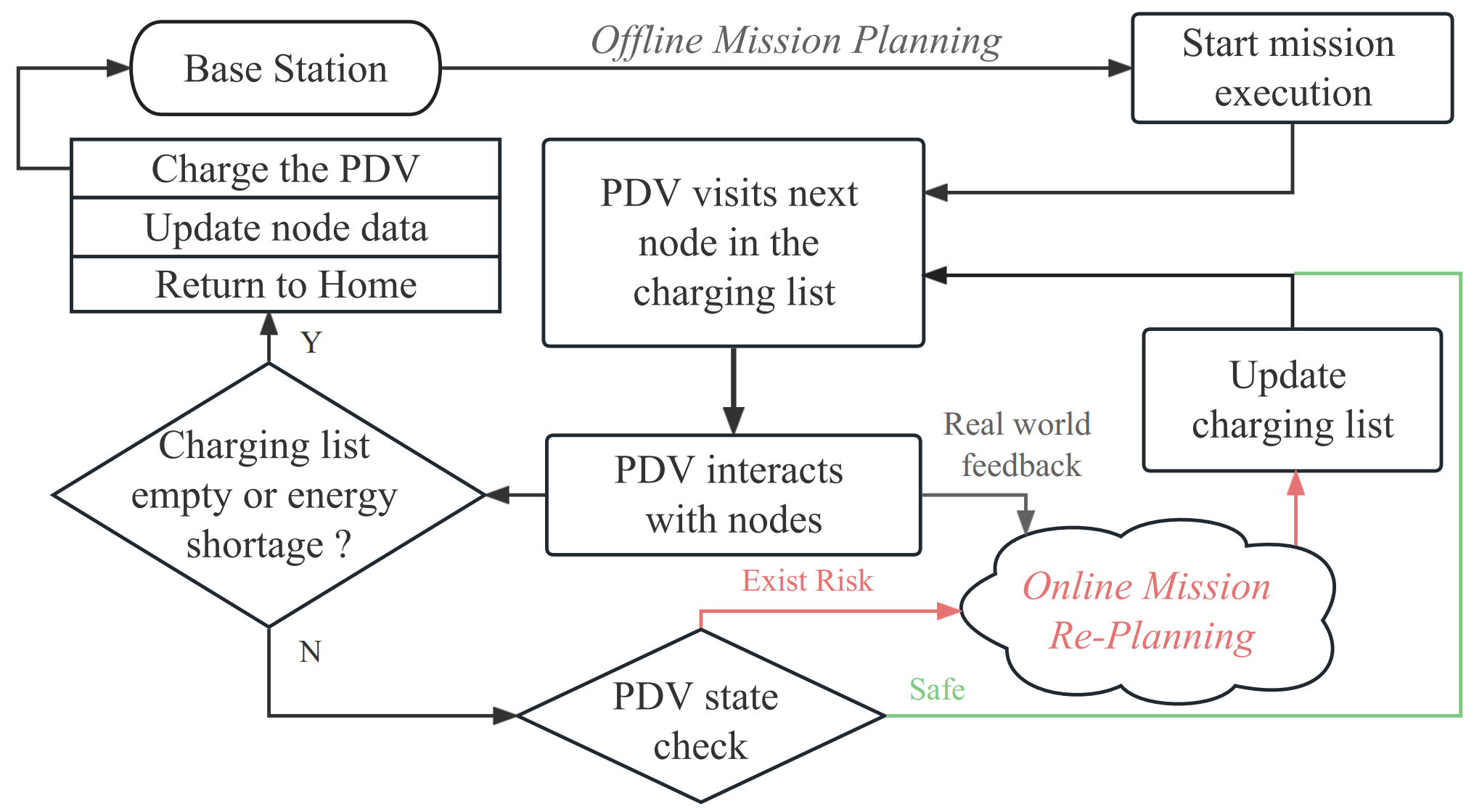}
\caption{System workflow}
\label{fig:sys}
\end{figure}
We introduce a two-stage, offline and online, mission planning scheme to tackle the CSP. The workflow is visualized in Fig.~\ref{fig:sys}. For a specific number of charging requests, the base station first conducts an offline mission planning phase, allocating a single UAV to complete the mission autonomously. There are various methods to make the system continuously autonomous. For example, a UAV can be deployed to collect the initial energy distribution of the WRSN and the energy decay rate of each sensor node. The UAV may conduct a routine online mission planning, utilizing the time taken for charging a sensor node to verify that it can complete its mission. Alternatively, if the UAV's remaining energy level is approaching a safety threshold value, it may also trigger the online scheme.
\begin{figure}
    \centering
    \begin{subfigure}{0.48\textwidth}
        \includegraphics[width=\textwidth]{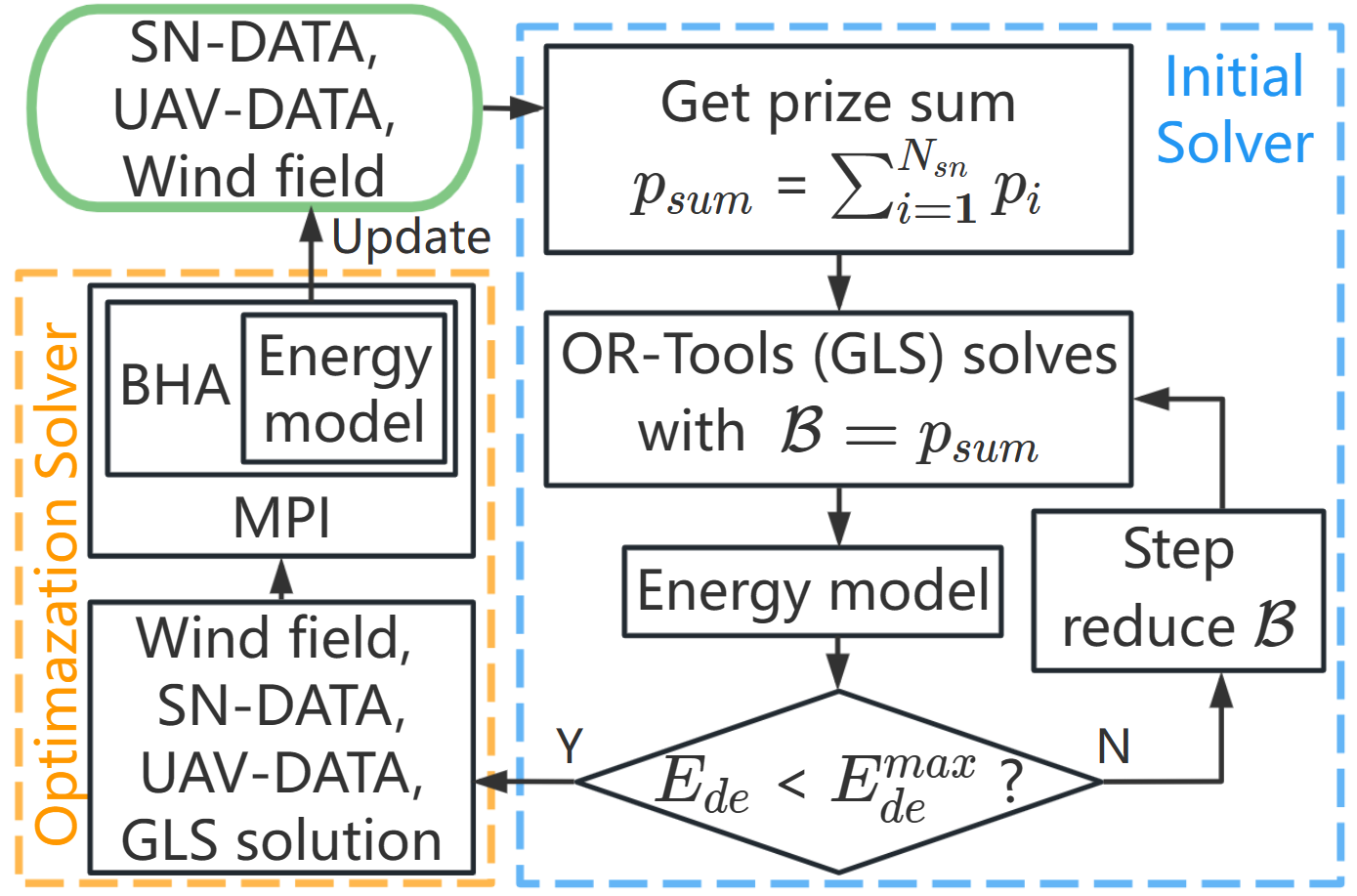}
        \subcaption{Architecture of ROMP, where OR-Tools (GLS) is the Google OR-Tools with Guided Local Search algorithm and $\mathcal{B}$ refers to the prize budget.}
        \label{fig:offline-scheme}
    \end{subfigure}
    \begin{subfigure}{0.48\textwidth}
        \centering
        \includegraphics[width=0.85\textwidth]{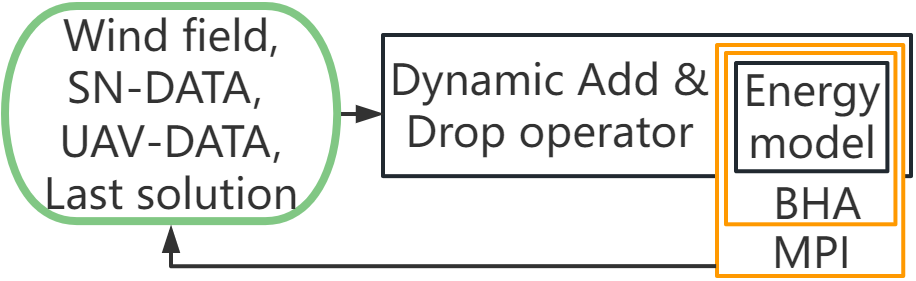}
        \subcaption{Architecture of FastROMP, where the last solution comes from the previous simulation.}
        \label{fig:online-scheme}
    \end{subfigure}
    \caption{Two different modes of ROMP, where SN-DATA includes sensor node deployment and their attributes; UAV-DATA includes the UAV coordinate, mission end coordinate and UAV remaining battery.}
\end{figure}

\subsection{Rapid Online Metaheuristic-based Planner (ROMP)}
\subsubsection{ROMP for Offline Mission Planning Scheme}
The base station, which is assumed to have adequate computing resources, conducts the offline scheme before the mission starts to assign a suitable charging plan for the UAV. The offline ROMP consists of an `Initial Solver' and an `Optimization Solver', whose architecture is visualized in Fig.~\ref{fig:offline-scheme}. The planner inputs are Sensor Node Data (SN-DATA) which includes sensor node coordinates, sensor type, capacitor voltage level and associated prize, and UAV Data (UAV-DATA) which includes UAV's present coordinate, mission end coordinate and the UAV's remaining battery. Because Google OR-Tools~\cite{ortools} is a mature commercial solver, we select it as the foundation of the `Initial Solver' that generates an initial solution comprising a maximum number of high-prize sensor nodes. In OR-Tools, we select GLS as the benchmark algorithm (see \S~\ref{sec:or-tools-benchmark}). It utilizes a `penalty' mechanism to drop sensor nodes, where a sensor node's `penalty' is proportional to its associated prize. Firstly, data from all sensor nodes and the UAV status will be gathered to obtain the prize sum $p_{sum}$. Because the UAV has limited charging capability, we assign a prize budget $\mathcal{B}$ to control the length of the charging list. Wind, sensor node deployment and UAV performance can significantly affect the appropriate budget value. As a result, the ‘Initial Solver’ must start with $\mathcal{B} = p_{sum}$ (i.e., the TSP scenario) and iteratively reduce $\mathcal{B}$ until the first feasible solution is found. Note that the step reduction loop is controlled by an individual process, namely speedup strategy, which balances algorithm convergence and computation time. The speedup coefficient is computed from the difference between simulated discharged energy and the given budget. For example, we conservatively reduce some prize budgets to skip unnecessary simulations if the difference exceeds a predetermined threshold.

The `Optimization Solver' further optimizes the initial solution in the given dynamic simulation environment. After $N_{gen}$ generations of evolution, where $N_{gen}$ is user-defined, the `Optimization Solver' will return the best-known solution to the UAV. We present the main algorithm, Context-aware Black Hole Algorithm (CBHA), in the `Optimization Solver' as follows. The BHA, first presented in~\cite{hatamlou2018bh}, is a metaheuristic optimization algorithm inspired by the black hole phenomenon. In BHA, the solutions to the optimization problem are represented as particles in a search space, and the best available solution is characterized as the black hole. The black hole has a probability $P_a$ to attract other solutions towards itself, causing stochastic displacements in the search space. CBHA mainly comprises three phases: initialization, attraction, and absorption.

\begin{algorithm}[!b]
    \SetKwInOut{Input}{Input}
    \SetKwInOut{Output}{Output}
    \Input{Sensor node set $V$, \textit{route} $\{ v_1, ..., v_N \}$,\\ target node $v_i$ in \textit{route}, search number $N_{srch}$}
    \Output{A candidate list of length $N_{srch}$}
    \caption{Candidate search operator}
    \label{alg:candidate-search}
    Record previous node $v_{i-1}$ and next node $v_{i+1}$ of target node $v_i$ in \textit{route}\;
    \For{node $v_k \in V \setminus \{ v_1, v_N, v_{i-1}, v_{i+1} \} $}{
        Compute fitness metric $M(v_k)$ (Eq.~\eqref{eq:dist-prize})\; 
        Append the pair $(M(v_k), v_k)$ to a list $L$\;
    }
    Order $L$ with descending order\;
    $\forall$ pair $(M(v_k), v_k) \in L$, sequentially add $v_k$ into candidate list until its length equals to $N_{srch}$\;
    \Return Candidate list.
\end{algorithm} 

The initialization phase generates a set of populations from an initial route, utilizing a probabilistic candidate search operator (see Alg.~\ref{alg:candidate-search}). This operator is typically more efficient than the original version because it considers the influence of index permutation on the fitness metric. It is applied to each element in the initial route. Moreover, to filter out some visit that has a low prize or that are too far from the black hole route, we employ a simplified fitness metric $M$ that is based on the fixed distance between sensor nodes and their corresponding prizes:
\begin{equation}
    \label{eq:dist-prize}
\begin{split}
    M & (v_j) = W_{re} \cdot \bigg(w(v_j) - lb\bigg) \\
    & -W_{de}\cdot lg\bigg(d(v_{i-1}, v_j) + d(v_j, v_{i+1})\bigg)
\end{split}
\end{equation}
where $W_{re}$ and $W_{de}$ are two weight coefficients for recharged and discharged energy separately; $d(v_{i-1}, v_j)$ (or $d(v_j, v_{i+1})$) is the distance between the senor node $v_j$ and the previous (or next) sensor node of $v_i$ in the target solution $S_t$. We subtracted $w(v_j)$ by $lb$ and applied a common logarithm to restrict these two factors within a similar magnitude scale because the prize ranges from $lb$ to $ub$, and distance ranges from meters to thousands of meters. CBHA is designed to prioritize the search for potential sensor nodes with higher prizes, resulting in lower priority assigned to sensor nodes around $lb$.

\begin{algorithm}[!t]
    \caption{Attraction phase}
    \label{alg:attraction}
    \SetKwInOut{Input}{Input}
    \SetKwInOut{Output}{Output}
    \Input{\textit{Route} $\{ v_1, ..., v_i, ..., v_N \}$, \\black hole $\textit{route}^{BH} \{ v_1, ..., v_i^{BH}, ..., v_N \}$}
    \Output{A modified \textit{route} after attraction phase}
    \For{$i = 2$ to $N_{route} - 1$}{
        \If{random [0, 1] $\leq P_a$ \textbf{and} $v_i \neq v_{i}^{BH}$}{
            Compute the new coordinate $\mathbf{x}_{i}^\prime$, using $v_i$ and $v_i^{BH}$ (Eq.~\eqref{eq:attraction})\;
            For $\mathbf{x}_{i}^\prime$, find a closest node $v_i^\prime \in \mathcal{V} \setminus \{ v_1, v_N \}$\;
            If $v_i^\prime \in \textit{route}$, swap $v_i$ and $v_i^\prime$\; Otherwise, $v_i \leftarrow v_i^\prime$\;
        }
    }
    \Return New \textit{route}.
\end{algorithm}
The attraction phase (Alg.~\ref{alg:attraction}) facilitates search space exploration by attracting candidate routes towards promising regions. The attraction of a node at index $i$ of a candidate route is formulated as:
\begin{equation}
\label{eq:attraction}
    \mathbf{x}_i^\prime = \mathbf{x}_i + \alpha \cdot (\mathbf{x}_i^{BH} - \mathbf{x}_i), \quad \forall i \in 2, ..., N_{route} - 1
\end{equation}
where $\mathbf{x}_i$ and $\mathbf{x}_i^{BH}$ are the coordinates of the node at index $i$ of a candidate route and a black hole, respectively. We then approximate the new coordinate $\mathbf{x}_i^\prime$ to the closest node in the feasible sensor node set $\mathcal{V}$ to complete this process.

\begin{algorithm}[!t]
    \caption{ROMP for online mission planning}
    \label{alg:online-romp}
    \SetKwInOut{Input}{Input}
    \SetKwInOut{Output}{Output}
    \Input{SN-DATA, UAV-DATA, wind field}
    \Output{Black hole $\textit{route}^{BH}$ \{$v_1, ..., v_i^{BH}, ..., v_N$\}}
    GLS solves with full prize budget $\mathcal{B} = p_{sum}$\;
    Obtain the initial route and evaluate its energy cost\;
    Execute the step reduction loop to obtain the first feasible \textit{route} $ \{v_1, ..., v_N\}$\;
    Initialize $N_{pop}$ populations from \textit{route} and update their fitness metric values (Eq.~\ref{eq:re-de})\;
    \While{\textbf{not} reach stopping criteria}{
        \For{route $\in$ populations}{
            Perform attraction phase (Alg.~\ref{alg:attraction}) to \textit{route}\;
            Compute the fitness metric of \textit{route}\;
            \If{new fitness metric is better}{
                Update new \textit{route} and its fitness metric\;
            }
        }
        Update black hole $\textit{route}^{BH}$ and its fitness metric\;
        Compute event horizon radius $r$\;
        \For{any population fitness metric $< r$}{
            Perform absorption phase to population \textit{route}\;
            Update new \textit{route} and its fitness metric\;
        }
        Update black hole $\textit{route}^{BH}$ and its fitness metric\;
    }
    \Return Black hole $\textit{route}^{BH}$.
\end{algorithm}

To maintain the diversity of the population and avoid stagnation at local extremes, the absorption phase re-initializes (same as the initialization phase) the population solution whose fitness metric is within the event horizon radius $r$. Moreover, to align with the context of CSP, we formulate the fitness metric of the entire route as follows:
\begin{equation}
\label{eq:re-de}
    \begin{split}
    M = \text{penalty}&~+~W_{re} \cdot \frac{\sum_{i=1}^{N_{route}}E_{re}}{\sum_{i=1}^{N}E_{re}} \\
    -& W_{de} \cdot \frac{\sum_{i=1}^{N_{route}} E_{IPT} + \int_{t_{v_1}}^{t_{v_N}} P_f \:dt}{E_{UAV}(t=t_{v_1})}
    \end{split}
\end{equation}
where the penalty is $-100$ when the route cannot satisfy Constraint \eqref{con:budget}, otherwise the penalty is $0$; the second term refers to the ratio of UAV recharged energy to all rechargeable energy of the WRSN; the third term denotes the ratio of UAV discharged energy to UAV initial energy.

Parallelization of serial CBHA can be achieved with Message Passing Interface (MPI) because population solutions are generated from a common source, enabling uniform initialization and evolution process distribution. Each process's black hole route is collected periodically (determined by aggregate frequency) to update a global black hole route. 

\subsubsection{ROMP for Online Mission Planning Scheme}
With minor modifications, ROMP can be employed for online mission planning. We present the pseudo-code of online ROMP as Alg.~\ref{alg:online-romp}. To assess the performance of ROMP, a comprehensive analysis is conducted in \S~\ref{sec:dync}. Our results indicate that, in specific large-scale scenarios, the step reduction loop in the ‘Initial Solver’ can be highly time-consuming. This means ROMP may exhibit limited performance when applied to online (re-)planning scenarios.

\subsection{FastROMP}
To further decrease the computational complexity of ROMP, we have developed FastROMP. This approach incorporates online updates such as UAV flight time, remaining battery level and mission progress into the planning process. This allows for dynamic adjustments to be made to the previously established route. Drawing inspiration from~\cite{kobeaga2018efficient}, we have designed an evaluation metric for dropping or adding a sensor node within the context of the CSP. The architecture of FastROMP is shown in Fig.~\ref{fig:online-scheme}. FastROMP begins by calibrating the route yielded from the previous computation to ensure its accuracy. If the solution cannot comply with the energy budget constraint, ROMP invokes the drop operator iteratively until a feasible solution is reached.  We define a metric, drop value, to evaluate the behaviour of dropping one sensor node $v_i$ (if removed at route index $i$):
\begin{equation}
\label{eq:drop}
    drop(v_i) = \frac{p(v_i)}{\int_{t_{v_{i-1}}}^{t_{v_i}} P_f \:dt + \int_{t_{v_i}}^{t_{v_{i+1}}} P_f \:dt - \int_{t_{v_{i-1}}}^{t_{v_{i+1}}} P_f \:dt }
\end{equation}
where the denominator term refers to the discharged energy difference if removed node $v_i$ from the route. $v_{i-1}$ and $v_{i+1}$ denote the previous node and next node in the route, respectively. Alg.~\ref{alg:drop} demonstrates the drop operator algorithm. Subsequently, the add operator is repeatedly applied until the solution attains maximal utilization of the allocated energy budget. Specifically, the add operator (Alg.~\ref{alg:add}) finds the best position for inserting the unvisited node. We define the add value of a sensor node $v_i$ (if inserted at route index $i$) as:
\begin{equation}
    \label{eq:add}
    add(v_i) = \begin{cases}
        \text{Eq.~\eqref{eq:drop}} & \text{satisfy Constraint~\eqref{con:budget} if added}\\
        0 & \text{otherwise}
    \end{cases}
\end{equation}
We used the same metric formulated for the drop value when the route is feasible after insertion. However, to guarantee the feasibility of the solution for further optimization, it is necessary to re-simulate the visit of any newly added/dropped sensor node at route index $i$ and all subsequent visits. To better align with the context of the dynamic CSP, we assign two lists to keep track of the energy consumption and the departure time for each visit.

\begin{algorithm}[!t]
    \caption{Dynamic Drop operator}
    \label{alg:drop}
    \SetKwInOut{Input}{Input}
    \SetKwInOut{Output}{Output}
    \Input{\textit{route} $\{ v_1, ..., v_N \}$, UAV energy list $\{ E_1, ..., E_N \}$, flight time list $\{ T_1, ..., T_N \}$}
    \Output{A feasible \textit{route} satisfies Constraint~\eqref{con:budget}}
    \While{Route does \textbf{not} satisfy Constraint~\eqref{con:budget}}{
        $\forall\:\text{node}\: v_i \in \textit{route}$, compute drop value (Eq.~\eqref{eq:drop})\;
        Order the drop value list and record the index of the lowest drop value as $i$\;
        Record $E_{i-1}$ and $T_{i-1}$, remove $v_i$ from \textit{route}\;
        Update UAV energy list and flight time list at \textit{route} index $i$ and after \textit{route} index $i$\;
    }
    \Return New \textit{route}, energy list and flight time list.
\end{algorithm}
\begin{algorithm}[!t]
    \caption{Dynamic Add operator}
    \label{alg:add}
    \SetKwInOut{Input}{Input}
    \SetKwInOut{Output}{Output}
    \Input{\textit{Route} $\{ v_1, ..., v_N \}$, UAV energy list $\{ E_1, ..., E_N \}$, flight time list $\{ T_1, ..., T_N \}$}
    \Output{A feasible \textit{route} satisfies Constraint~\eqref{con:budget}}
    \While{nodes can be inserted into route}{
        \For{node $v \notin$ route}{
            Get 3 neighbours in \textit{route} with minimum cost to visit $v$, $NBR = \{ nbr_1, nbr_2, nbr_3 \}$\;
            $\forall$ pair $\in NBR$ are adjacent in \textit{route}, find the pair minimizes the visit cost\;
            \If{\textbf{no} adjacent pair exists}{
                $\forall nbr \in NBR$, find its previous and next node in \textit{route}, create pair $(v_{prev}, nbr)$ or $(nbr, v_{next})$ minimizes the visit cost\;
            }
            Compute add value of node $v$ (Eq.~\eqref{eq:add}) and append it into add value list $L$\;
        }
        Order $L$ and record the node with the highest positive add value as $v_i$ and its insertion index in \textit{route} as $i$\;
        Record $E_{i-1}$ and $T_{i-1}$, insert $v_i$ into \textit{route}\;
        Update UAV energy list and flight time list at \textit{route} index $i$ and after \textit{route} index $i$\;
    }
    \Return New \textit{route}, energy list and flight time list.
\end{algorithm}
\begin{algorithm}[!t]
    \caption{FastROMP}
    \label{alg:fastromp}
    \SetKwInOut{Input}{Input}
    \SetKwInOut{Output}{Output}
    \Input{Previous \textit{route} \{$v_1, ..., v_i, ..., v_{N}$\},
    SN-DATA, UAV-DATA, wind field}
    \Output{Black hole $\textit{route}^{BH}$ \{$v_1, ..., v_i^{BH}, ..., v_N$\}}
    Calibrate previous \textit{route}, update UAV energy list and flight time list\;
    Apply drop operator (Alg.~\ref{alg:drop}) and add operator (Alg.~\ref{alg:add}) to previous \textit{route}\;
    Initialize $N_{pop}$ populations from adjusted \textit{route} and update their fitness metric values (Eq.~\eqref{eq:re-de})\;
    Apply 2-opt local search to each population\;
    Update black hole $\textit{route}^{BH}$ and its fitness metric\;
    \While{\textbf{not} reach stopping criteria}{
        \For{route $\in$ populations}{
            Perform attraction phase (Alg.~\ref{alg:attraction}) and 2-opt local search to \textit{route}\;
            Compute the fitness metric of \textit{route}\;
            \If{new fitness metric is better}{
                Update new \textit{route} and its fitness metric\;
            }
        }
        Update black hole $\textit{route}^{BH}$ and its fitness metric\;
        Compute event horizon radius $r$\;
        \For{any population fitness metric $< r$}{
            Perform absorption phase and 2-opt local search to the population's \textit{route}\;
            Update new \textit{route} and its fitness metric\;
        }
        Update black hole $\textit{route}^{BH}$ and its fitness metric\;
    }
    Apply drop operator and add operator to $\textit{route}^{BH}$, update UAV energy list and flight time list\;
    \Return Black hole $\textit{route}^{BH}$.
\end{algorithm}

Alg.~\ref{alg:fastromp} shows the pseudo-code of FastROMP. Since the greedy drop and add operators can drive the previous route towards local optimum, we use CBHA to explore the potential for finding a global optimal route. CBHA's stopping criteria can be determined by one or more factors, such as the number of iterations performed, execution time elapsed, or the attainment of a target fitness metric. FastROMP invokes drop and add operators again at its end to ensure the route's feasibility and the energy budget's maximal utilization.

\section{Experiments, Results and Discussion}
\begin{figure*}[!t]
\sbox\twosubbox{%
  \resizebox{\dimexpr0.99\textwidth-1em}{!}{%
    \includegraphics[height=4cm]{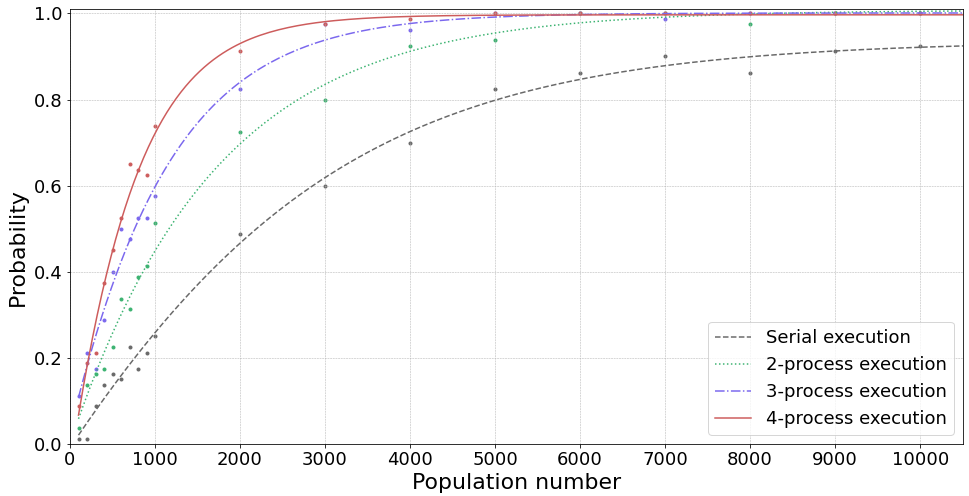}%
    \includegraphics[height=4cm]{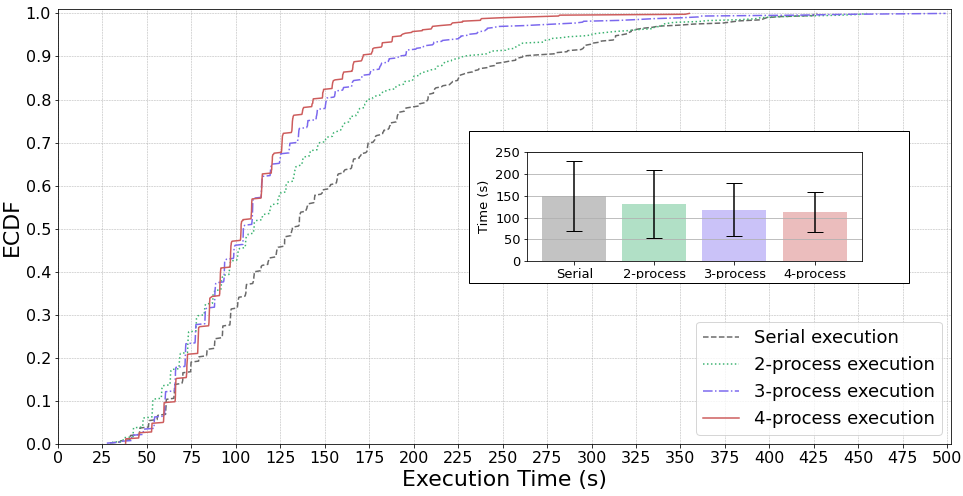}%
  }%
}
\setlength{\twosubht}{\ht\twosubbox}
\centering
\subcaptionbox{Parallelization performance summary. The probability is the percentage that CBHA obtains a solution better than GLS over all executions. The 4-process execution is capable of achieving 100\% convergence at 5000 populations, while the serial execution fails to converge even with a population size of 10000.\label{fig:para1}}{%
  \includegraphics[height=\twosubht]{figure/para_gen100.png}%
}\quad
\subcaptionbox{Empirical cumulative distribution function for finding better solutions against time using ROMP. The results span 500 executions using $N_{pop} = 5000$ and $N_{gen} = 100$. The 4-process execution can save $\sim$30\% execution time over serial while achieving better solution quality.\label{fig:para2}}{%
  \includegraphics[height=\twosubht]{figure/para_pop5000_gen100.png}%
}
\caption{Parallel search performance analysis}
\end{figure*}

This section presents numerical results to evaluate offline and online mission planning schemes. The planner was coded in C++20 and executed on a 64-bit Intel Core i7-8550U CPU running at 1.80 GHz with 8 GB RAM (i.e., DJI Manifold 2-C~\cite{Manifold}, a typical onboard flight computer). The following attributes (Attrs) were considered when evaluating the solution performance:
\begin{itemize}
    \item[$t$] Algorithm execution time [unit: s]
    \item[$E_{re}^{\star}$] Recharged energy of visited sensor nodes in a single mission [unit: J]
    \item[$E_{de}^{\star}$] UAV discharged energy in a single mission [unit: Wh]
    \item[$R_{re}$] The ratio of total recharged energy to all rechargeable energy of requested sensor nodes [unit: \%]
    \item[$R_{de}$] The ratio of total UAV discharged energy to UAV initial energy [unit: \%]
    \item[$R_{rd}$] The conversion efficiency from the UAV's discharged energy to sensor node recharged energy [unit: \textperthousand] 
\end{itemize}
where:
\begin{align}
    R_{re} &= \Big(E_{re}^{\star}/\sum\nolimits_{i=2}^{N - 1} E_{re}(v_i)\Big)~\cdot~100\%\\
    R_{de} &= \Big(E_{de}^{\star} / \mathcal B\Big)~\cdot~100\%\\
    R_{rd} &= \Big(E_{re}^\star / (3600 \cdot E_{de}^\star)\Big)~\cdot~1000\text{\textperthousand}
\end{align}

\begin{figure}[!b]
\centering
\includegraphics[width=0.486\textwidth]{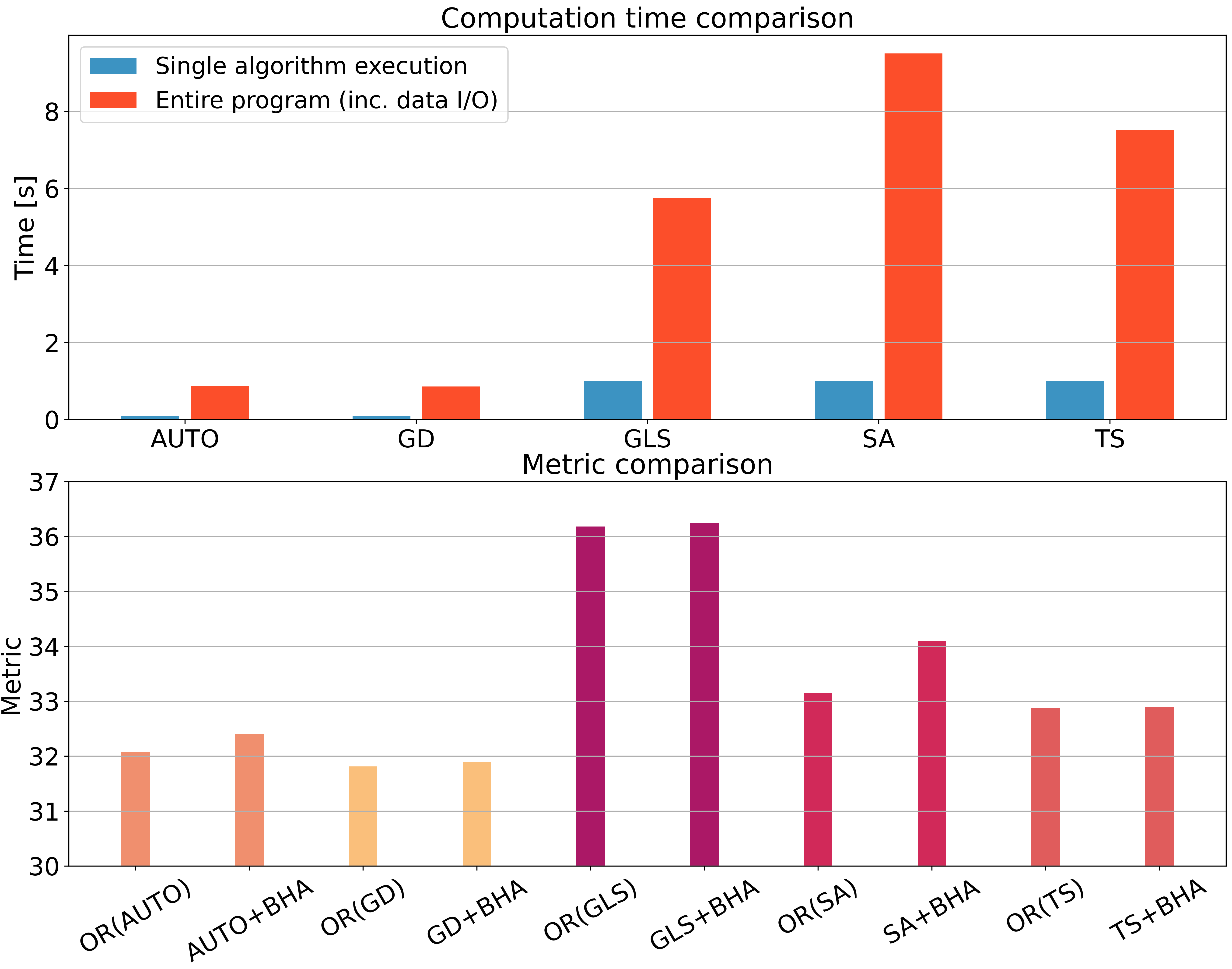}
\caption{Execution results of OR-Tools and the `Optimization Solver' with different local search options, including Automatic selection (AUTO), Greedy Descent (GD), GLS, Simulated Annealing (SA), and Tabu Search (TS). GLS was chosen as the benchmark algorithm, given the favourable balance of computational efficiency and solution quality.}
\label{fig:ortest}
\end{figure}

\begin{figure*}[!htbp]
    \centering
    \includegraphics[width=\textwidth, height=0.34\textwidth]{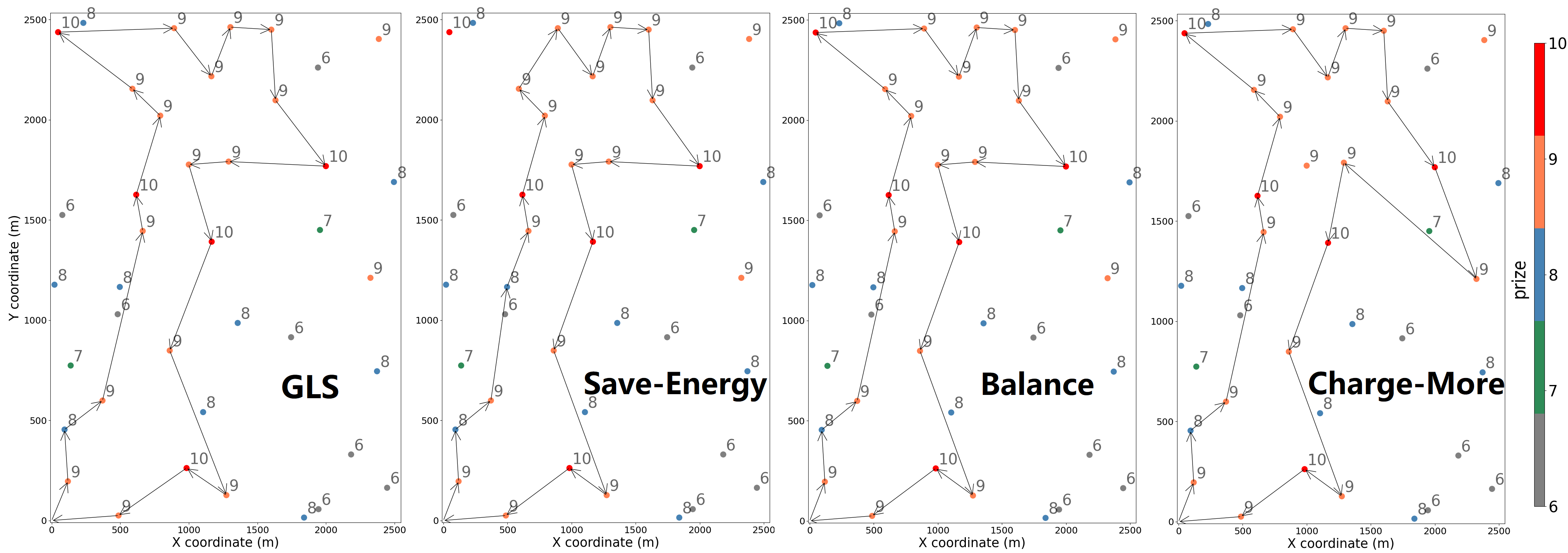}
    \caption{An OP-1 instance of GLS' and ROMP's solutions using different search strategies.}
    \label{fig:typ}
\end{figure*}

We set the lower prize bound $lb$ to 6 and upper bound $ub$ to 10 to distinguish between relatively discharged (6-10) and charged (1-5) sensor nodes and to reduce the number of sensor nodes involved in the computation. Arguments in the algorithm can be tuned according to different requirements. Setting a large population and generation number can, for example, lead to deep searches and long computation time. In CBHA, we set the population number $N_{pop}$ to 80, generation number $N_{gen}$ to 80, `attraction' probability $P_a$ to 75\%, search number $N_{srch}$ to 10, event horizon radius $r$ at 25\%, and aggregate frequency as 10 as a common setting. The parameter settings for $P_a$, $N_{srch}$ and $r$ are selected to allow the algorithm to evolve more actively. It is important to note that updating the global solution should not frequently occur since this process involves collecting routes from other processes, sorting fitness metrics, broadcasting a global route and updating the black hole route, which can have a non-trivial impact on the execution time. The update frequency represents a trade-off between computing time and search efficiency. These chosen values were based on a scenario where a UAV requires tens of seconds to charge a supercapacitor. Thus, ROMP will produce a safe route, but it may not be more efficient than GLS. A detailed analysis of $N_{pop}$ and $N_{gen}$ setting can be seen in \S~\ref{sec:parallel-search-experiment}, which shows that setting the population to 5000 allows our algorithm to produce a better solution than GLS in 100\% of cases tested.  

\subsection{Meta-heuristic Algorithm Options in OR-Tools}
\label{sec:or-tools-benchmark}
OR-Tools supports different local search options: AUTO, GD, GLS, SA, and TS~\cite{ortools}. These options were tested and averaged over 20 executions (see Fig.~\ref{fig:ortest}). AUTO and GD completed the calculation quickly but yielded solutions with the lowest fitness metric. Note that the solution provided by OR-Tools is a basis of CBHA's calculation, which means a low-quality initial solution will increase the opportunity for the CBHA to find a better solution, which is not helpful for performance analysis. Therefore, we selected GLS as the benchmark algorithm, as it found the solutions with the highest fitness metric in an acceptable computation time. 

\begin{figure}[!b]
    \centering
    \includegraphics[width=0.47\textwidth]{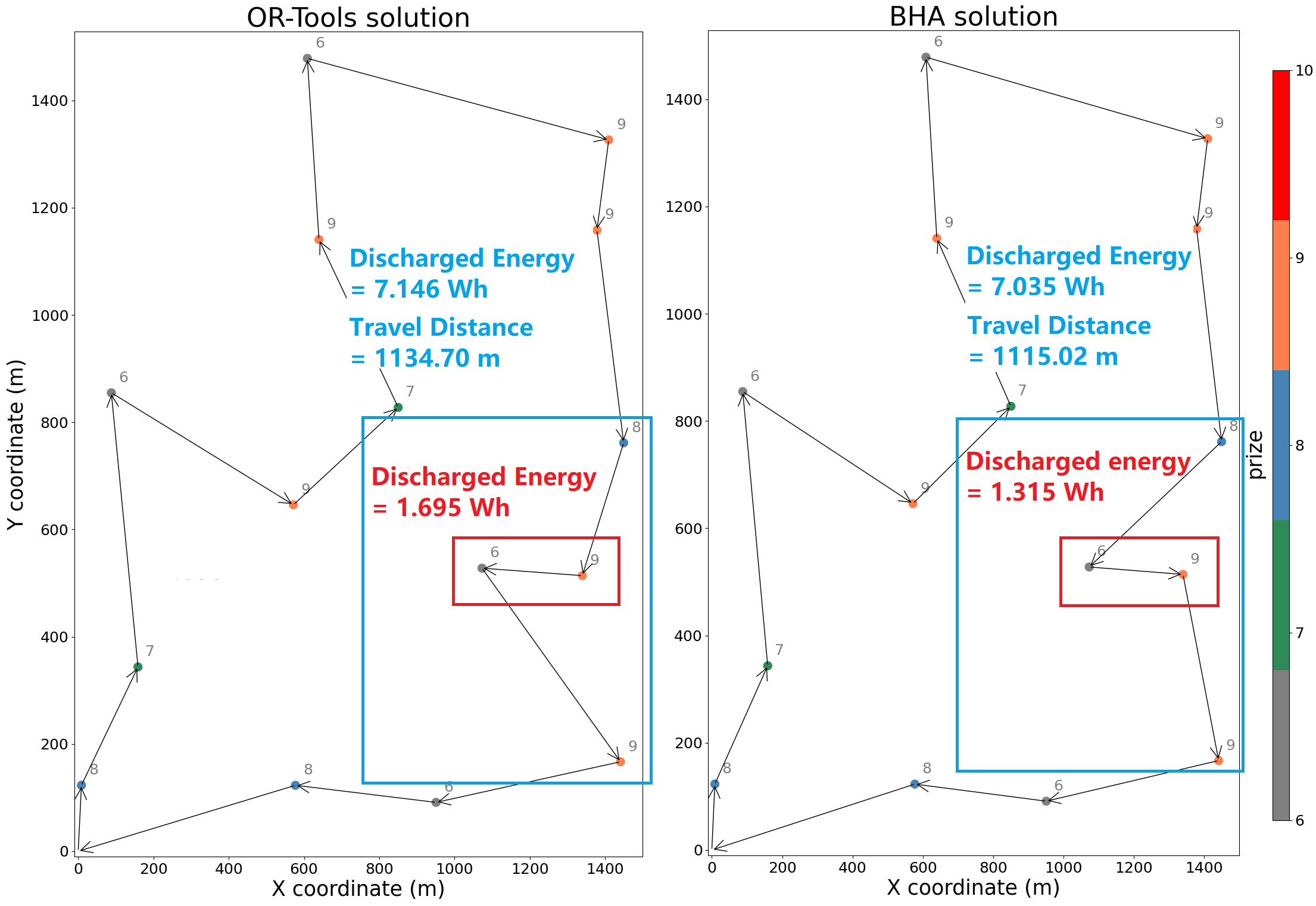}
    \caption{An instance showing changes in route sequence and discharged energy after applying a constant west wind.}
    \label{fig:wind_effect}
\end{figure}

\subsection{Parallel Search Efficiency of the CBHA}
\label{sec:parallel-search-experiment}
Parallelization performance is analyzed based on the OP scenario, which aims at obtaining the probability that the `Optimization Solver' finds a route with a better fitness value than the one from GLS, with varying numbers of processes over which computation can be distributed. Because CBHA is a population-based heuristic algorithm, the number of generations has little impact on performance. We therefore fixed the generation number to 100 for these experiments. Fig.~\ref{fig:para1} demonstrates the parallelization performance with constraints of increasing population numbers over 100 executions. A logistic function curve was fitted to the data points to show the trend with increasing populations. Results indicate that the serial execution's probability of finding a better solution cannot converge to 100\% even using 10000 populations, while the 4-processes execution converges to 100\% at 5000 populations. For a small population number, e.g., $N_{pop} \leq 500$, parallel execution achieves better solutions with approximately linear probability improvement. The execution time for the `optimization solver' to achieve a trusted solution can therefore be reduced linearly with increasing allocated processes. To find the minimum execution time needed to gain a better solution than GLS, another experiment was conducted using 5000 populations, and 100 generations, in which the 4-process execution can converge to 100\%. Fig.~\ref{fig:para2} demonstrates the empirical cumulative distribution (ECDF) of the execution time to achieve a better solution. With 4 allocated processes, the CBHA needs at least 5 minutes to guarantee a solution of better performance than OR-Tools. But compared to serial execution, the 4-process execution can save computation time by $\sim$30\% of the total execution time. Among all cases where CBHA achieved better solutions, using four processes had a smaller mean (112.04 s) and Standard Deviation (SD, 45.71) of execution time compared to serial execution (mean 149.76 s, SD 81.29). Although a simple strategy was used in this experiment, the results clearly show that parallelization can enhance the solver's stability and reduce the computational time needed for convergence. The experiment also proves that ROMP can be used for online re-planning on regular flight computers.

\begin{table}[!b]
\large
\centering
\caption{System performance in OP-1 and OP-2 scenarios}
\label{tab:tspop}
\begin{adjustbox}{width=0.486\textwidth}
\begin{tabular}{c||c||c||c||c||c||c||c}
\toprule
\multicolumn{2}{c||}{\diagbox{Attrs}{Cases}} & \multicolumn{3}{c||}{OP-1} & \multicolumn{3}{c}{OP-2} \\
\hline
\multirow{5}{*}{GLS} & $E_{re}^\star$ & \multicolumn{3}{c||}{540.71} & \multicolumn{3}{c}{224.77} \\
\cline{2-8}
& $R_{re}$ & \multicolumn{3}{c||}{59.382} & \multicolumn{3}{c}{52.721} \\
\cline{2-8}
& $E_{de}^\star$ & \multicolumn{3}{c||}{71.879} & \multicolumn{3}{c}{68.594} \\
\cline{2-8}
& $R_{de}$ & \multicolumn{3}{c||}{71.879} & \multicolumn{3}{c}{68.663} \\
\cline{2-8}
& $R_{rd}$ & \multicolumn{3}{c||}{2.0917} & \multicolumn{3}{c}{0.9102} \\
\midrule
\multirow{6}{*}{} & $W_{re}$ & 20 & 50 & 80 & 20 & 50 & 80 \\
\cline{2-8}
& $E_{re}^\star$ & 538.70 & 543.39 & 545.94 & 211.63 & 232.60 & 263.00\\
\cline{2-8}
& $R_{re}$ & 59.163 & 59.679 & 59.953 & 49.653 & 54.577 & 61.679 \\
\cline{2-8}
ROMP & $E_{de}^\star$ & 65.745 & 71.279 & 71.721 & 59.077 & 64.123 & 75.592 \\
\cline{2-8}
& $R_{de}$ & 65.810 & 71.350 & 71.793 & 59.137 & 64.187 & 75.668\\
\cline{2-8}
& $R_{rd}$ & {\textit{\textbf{2.2749}}} & \textbf{2.1174} & \textbf{2.1141} & \textbf{0.9915} & \textit{\textbf{1.0068}} & \textbf{0.9674} \\
\bottomrule
\end{tabular}
\end{adjustbox}
\end{table}

\subsection{Wind Impact on Solution Performance}
We selected the TSP scenario to explore the effect of wind on the `Optimization Solver' performance because additional energy consumption caused by wind (instead of other reasons like sequence order change) can be easily observed. For instance, in the last iteration of case 140 in Table.~\ref{tab:iter}, ROMP found a solution that saved 0.330 Wh whilst a north-east 3.606 m/s constant wind was present by reversing the visiting sequence of the initial solution. Another interesting comparison, where we tested 100 TSP cases with random sensor node deployments, is shown in Fig.~\ref{fig:wind_effect}, where we observe a 0.112 Wh reduction in discharged energy under a constant westerly 5 m/s wind. 

\subsection{ROMP Performance in OP Scenarios}
\label{sec:tspop}

\begin{figure*}[!htbp]
    \centering
    \includegraphics[width=\textwidth, height=0.34\textwidth]{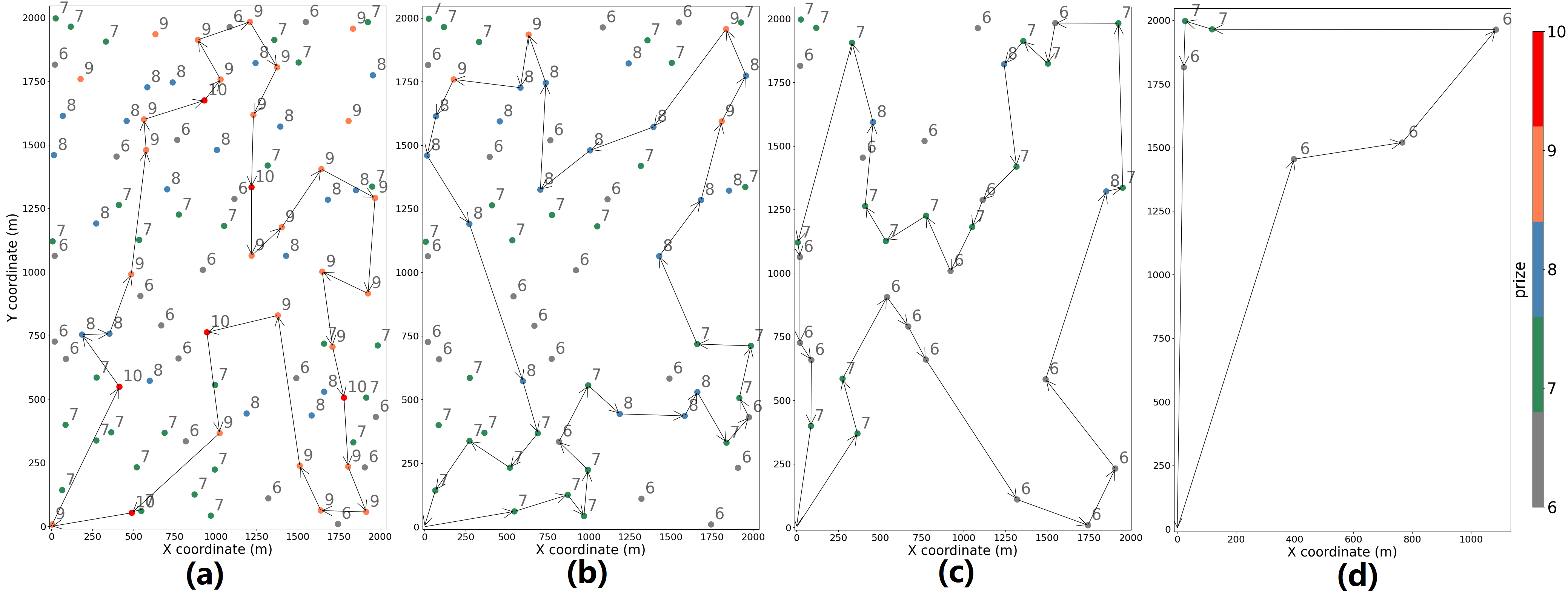}
    \caption{ROMP's solution for charging 100 sensor nodes: (a) to (d) sequentially illustrate the 4 required missions ($N_{iter}$) from first to last, respectively.}
    \label{fig:iter}
\end{figure*}
\begin{table*}[!t]
\centering
\large
\caption{System performance summary when increasing sensor node density from 10 to 150 over a 4 km$^2$ area.}
\label{tab:iter}
\begin{adjustbox}{width=\textwidth}
\begin{tabular}{c||c||c||c||c||c||c||c||c||c||c||c||c||c||c||c||c}
\toprule
\multicolumn{2}{c||}{\diagbox{Attrs}{$N_{sn}$}} & 10 & 20 & 30 & 40 & 50 & 60 & 70 & 80 & 90 & 100 & 110 & 120 & 130 & 140 & 150 \\ \hline
\multicolumn{2}{c||}{$N_{iter}$} & 1 & 1 & 2 & 2 & 2 & 2 & 3 & 3 & 3 & 4 & 4 & 4 & 4 & 5 & 5 \\
\hline
$E_{re}^*$ & avg. & 238.122 & 490.995 & 755.715 & 913.2545 & 1080.06 & 1332.19 & 1700.87 & 1890.48 & 2085.28 & 2317.19 & 2431.72 & 2863.22 & 2906.62 & 3045.98 & 3380.29 \\
\cline{2-17}
GLS & SD & 0 & 0 & 0.0004 & 7.1671 & 0.0003 & 0.0004 & 8.86092 & 3.62709 & 13.8216 & 11.2704 & 16.353 & 16.5165 & 26.946 & 43.6128 & 41.5022 \\
 \hline
$E_{re}^*$ & avg. & 238.122 & 490.995 & 755.715 & 915.788 & 1080.06 & 1332.19 & 1703.9 & 1891.31 & 2087.52 & 2319.01 & 2453.04 & 2865.89 & 2928.28 & 3023.82 & 3418.30 \\
\cline{2-17}
ROMP & SD & 0 & 0 & 0.0004 & 5.263 & 0.0003 & 0.0004 & 5.98282 & 0.0214 & 10.7482 & 7.16623 & 10.5066 & 11.5936 & 18.0121 & 20.5924 & 11.5158 \\
 \hline
$E_{de}^*$  & avg. & 47.881 & 63.487 & 106.523 & 127.98 & 142.177 & 145.875 & 189.267 & 208.13 & 220.066 & 264.007 & 272.738 & 285.547 & 300.461 & 340.141 & 375.619 \\
\cline{2-17}
GLS & SD & 0 & 0 & 1.7204 & 2.0782 & 1.0264 & 1.70823 & 4.67758 & 3.95527 & 4.34005 & 3.93993 & 5.10223 & 6.32977 & 18.0749 & 15.7338 & 15.139 \\
  \hline
$E_{de}^*$ & avg. & 47.881 & 63.487 & 106.523 & 128.036 & 142.177 & 145.861 & 188.883 & 207.312 & 219.261 & 263.474 & 271.58 & 284.397 & 297.316 & 328.489 & 355.54 \\
\cline{2-17}
ROMP & SD & 0 & 0 & 1.7204 & 2.0524 & 1.0264 & 1.68918 & 4.04203 & 1.94521 & 3.95358 & 3.79448 & 5.26239 & 7.275 & 17.4782 & 13.5864 & 11.6472 \\
\hline
\multicolumn{2}{c||}{$R_{rd}$ GLS} & 1.3814 & 2.1483 & 1.9707 & 1.9822 & 2.1102 & 2.5368 & 2.4963 & 2.5231 & 2.6321 & 2.4381 & 2.4767 & 2.7853 & 2.6871 & 2.4875 & 2.4997 \\
\hline
\multicolumn{2}{c||}{$R_{rd}$ ROMP} & 1.3814 & 2.1483 & 1.9707 & 1.9868 & 2.1102 & 2.5368 & 2.5058 & 2.5341 & 2.6446 & 2.4449 & 2.5090 & 2.7992 & 2.7358 & 2.4694 & 2.6706 \\
\bottomrule
\end{tabular}
\end{adjustbox}
\end{table*}

In this section, we focus on OP scenarios, given the acceptable solution quality of GLS for TSP scenarios. Two scenarios were tested in our experiments: OP-1, a scenario with many sensor nodes randomly deployed in a middle-scale area size (40 sensor nodes in a 2500 $\times$ 2500 m$^2$ grid), and OP-2, a scenario with few nodes randomly deployed in a large-scale area (20 sensor nodes in a 4000 $\times$ 4000 m$^2$ grid). Note that although the attribute of every sensor node is generated randomly in each case, the coordinates of all sensor nodes remain the same in each OP-1 and OP-2 experiment. This is to test the sensitivity of the planner to prize changes. We consider three search strategies: `Charge-More' ($W_{re} = 80$), `Balance' ($W_{re} = 50$), and `Save-Energy' ($W_{re} = 20$). Fig.~\ref{fig:typ} shows a visualization of typical results using different search strategies. For example, compared with the GLS's route, the `Charge-More' strategy replaces a temperature sensor node with prize 9 in the middle with another pressure sensor node with prize 9 in the middle right, which leads to an extra charging of 18.19 J energy.

To assess ROMP's performance in general scenarios, we demonstrate its averaged execution results in Table.~\ref{tab:tspop} among 300 randomized cases, where highlights in bold indicate \textbf{distinctly improved solutions}, relative to an initial solution and highlights in italics and bold indicate \textit{\textbf{the best solution}}. The search strategy's impact on results varies depending on the grid scale and sensor node density. For example, in OP-2, the `Charge-More' strategy resulted in $\sim$10\% more network energy charged than the `Save-Energy' strategy, while in OP-1, the difference is less than 1\%. An increasing trend of recharged energy ratio $R_{re}$ can be observed with increases of $W_{re}$ and $R_{de}$. Because of the biased penalty design, GLS tends to first add more sensor nodes with higher prize weight into the solution list. The solution length, discharged energy, and recharged energy may be limited because the collected prize reaches the threshold value of `capacity' quickly during searching. This can explain why UAV discharged energy is $\sim$70\% in most OP cases. 

\begin{figure*}[htbp]
    \includegraphics[width=\textwidth, height=0.34\textwidth]{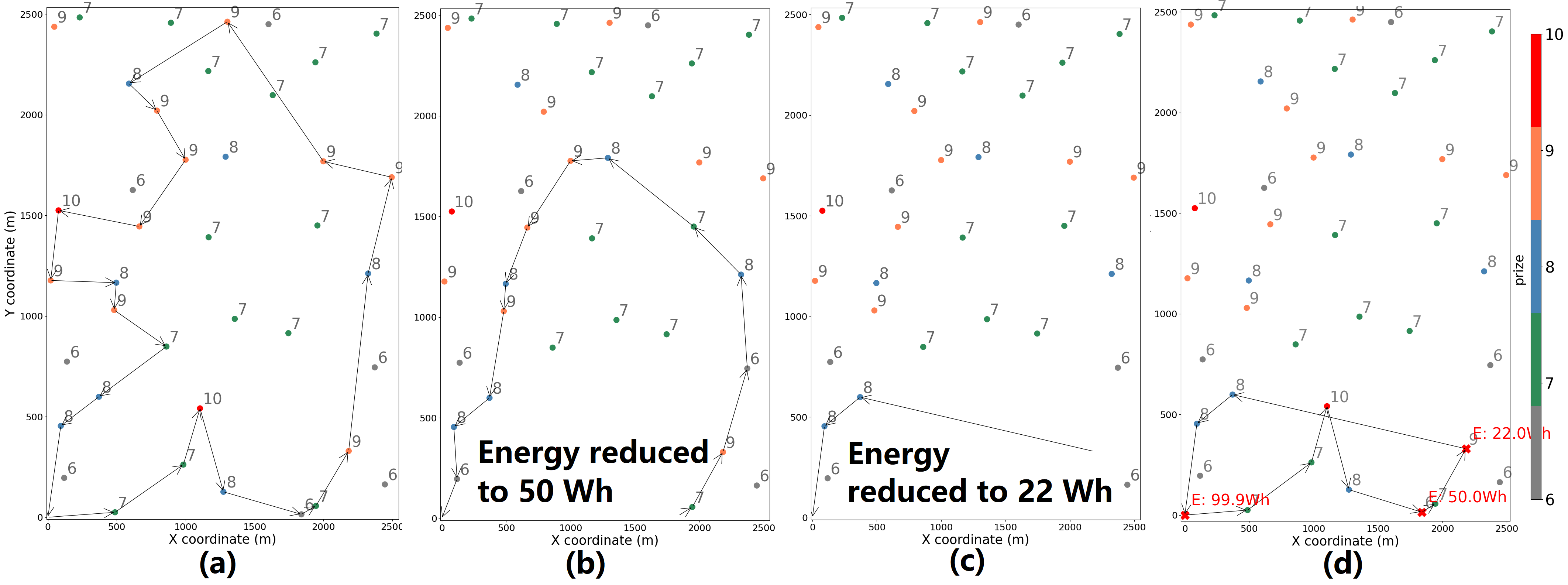}
    \caption{An instance of flight plans with dynamic energy update (initial energy 99.9 Wh), where (a) shows the original route of the given network; (b) shows the ROMP's route when UAV's remaining energy is reduced to 50 Wh; (c) shows the ROMP's route when UAV's remaining energy is reduced to 22 Wh; (d) is the actual route after re-planning of (b) and (c).}
    \label{fig:dyn}
\end{figure*}

\subsection{ROMP Performance with Increasing Nodes}
\label{sec:iter}
A functional requirement of our mission planner is to charge an entire network using a single UAV. A network consisting of 10 to 150 sensor nodes in a 2000 $\times$ 2000 m$^2$ grid is used to evaluate our system's performance, where the attribute of each sensor node is randomized. System performance is indicated by the number of UAV mission iterations $N_{iter}$ required to recharge the entire network and the attributes listed in Table.~\ref{tab:iter}. Note that in this experiment, we did not use metrics of $R_{re}$ and $R_{de}$ because the UAV may not have enough unvisited sensors in some missions, which leads to an unfair comparison of $R_{re}$ and $R_{de}$. All execution results are averaged over 20 executions. GLS performance degrades as network scale increases, which leaves more space for ROMP to improve. An example can be found in case 150, where the results show that ROMP successfully improves the solutions provided by GLS with an average of 38.01 J more recharged energy and 20.079 Wh less discharged energy in total. Fig.~\ref{fig:iter} shows the ROMP's routes to charge 100 sensor nodes, requiring four missions (Fig.~\ref{fig:iter}a to Fig.~\ref{fig:iter}d).

\subsection{System Performance with Dynamic Updates of the UAV Remaining Energy}
\label{sec:dync}
To simulate scenarios where a UAV encounters an unpredictable force, such as an unexpected headwind that causes its remaining energy to be depleted at a faster rate, we set the UAV's remaining energy to be arbitrarily reduced to a sufficiently low level whilst it is executing its route. Fig.~\ref{fig:dyn} shows how ROMP adjusts the UAV's route. The panels from left to right illustrate the original route, the first updated route after the UAV's remaining energy is reduced, the second updated route after a further reduction in remaining energy, and the final route. The first change occurs once the UAV has charged five sensor nodes, after which its remaining energy is reduced to 50 Wh. The second change occurs once the UAV has recharged two additional sensor nodes, after which its remaining energy is further reduced to 22 Wh. In this case, the UAV may not have enough energy to execute its original route. The planner, therefore, adjusts the estimated mission execution capability of the UAV, which yields a route of 16.18 Wh discharged energy.

\begin{figure}[!b]
    \centering
    \begin{subfigure}{0.486\textwidth}
        \includegraphics[width=\textwidth]{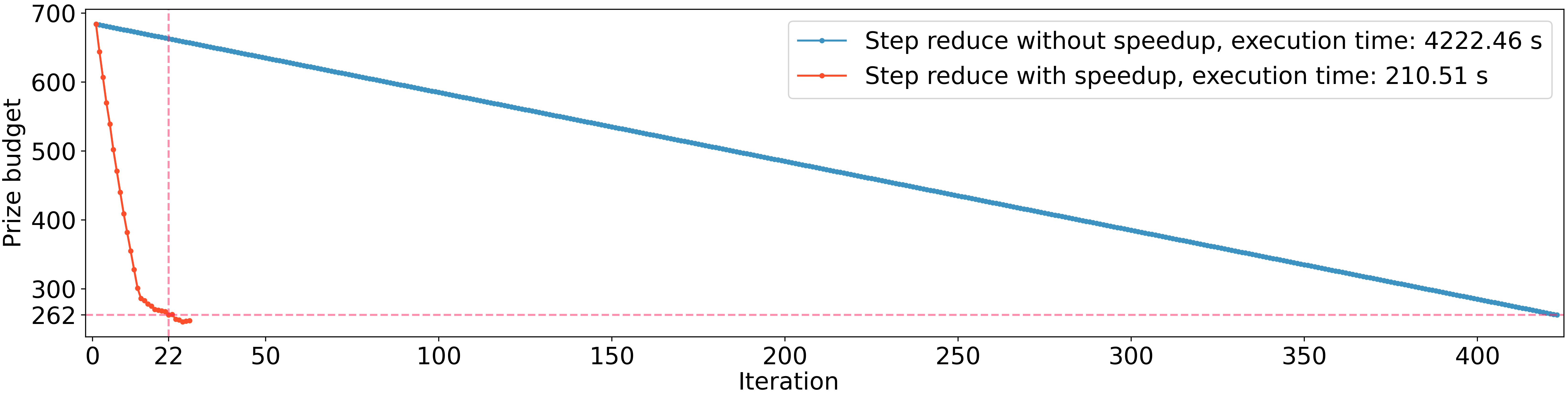}
        \subcaption{Convergence w.r.t. prize budget, where the red horizontal dashed line shows the first feasible solution found by both strategies; the red vertical dashed line denotes the iteration number needed for the step reduction loop with speedup strategy to find the first feasible solution.}
        \label{fig:sp-bgt}
    \end{subfigure}
    \begin{subfigure}{0.486\textwidth}
        \includegraphics[width=\textwidth]{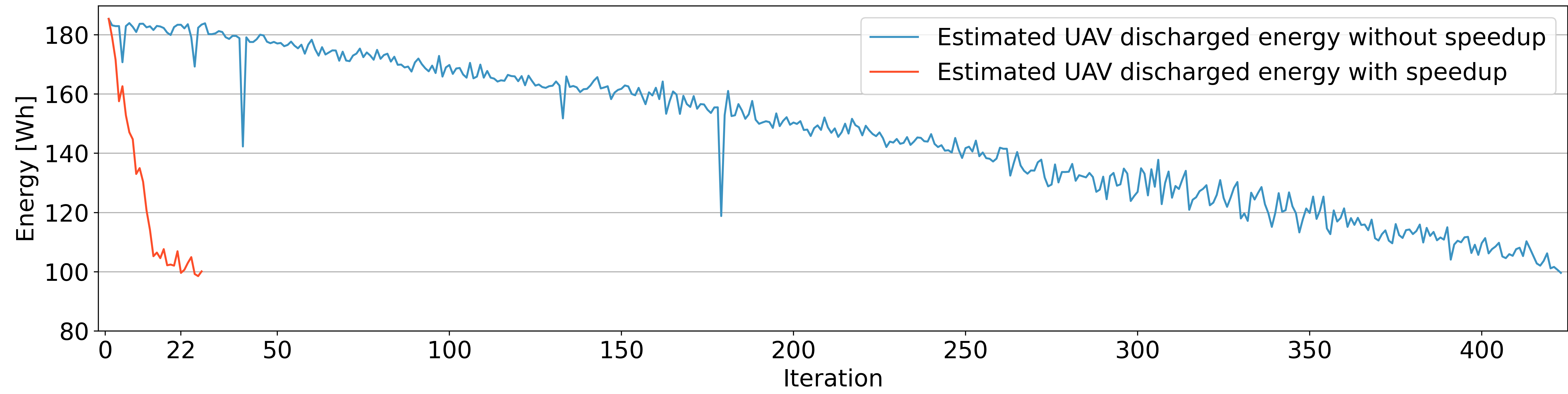}
        \subcaption{Convergence w.r.t. discharged energy, where the target energy level is 99.9 Wh.}
        \label{fig:sp-eng}
    \end{subfigure}
    \caption{The step reduction loop convergence process (with and without speedup strategy).}
\end{figure}
\begin{figure*}[htbp]
    \centering
    \begin{subfigure}{\textwidth}
        \includegraphics[width=\textwidth]{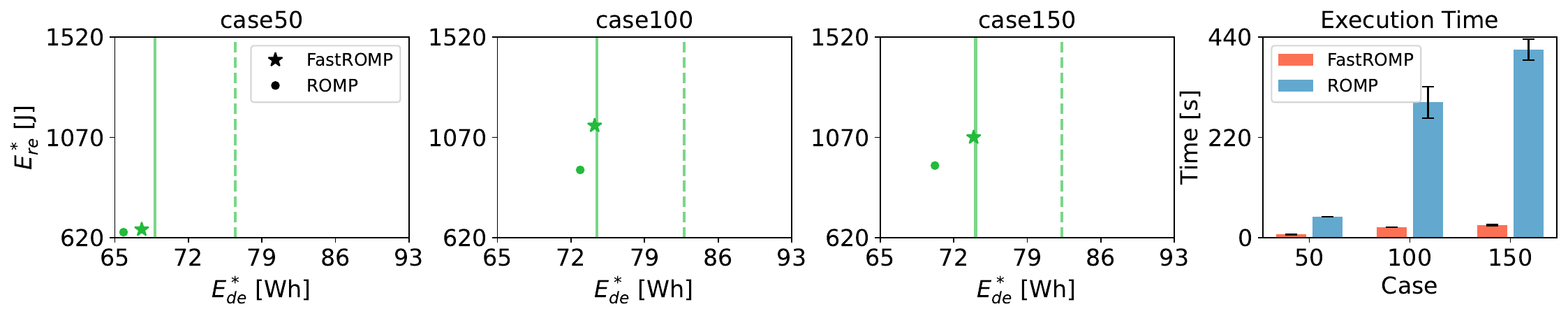}
        \caption{UAV energy budget is 90\% of the expected battery level.}
        \label{fig:dync-romp-top}
    \end{subfigure}
    \begin{subfigure}{\textwidth}
        \includegraphics[width=\textwidth]{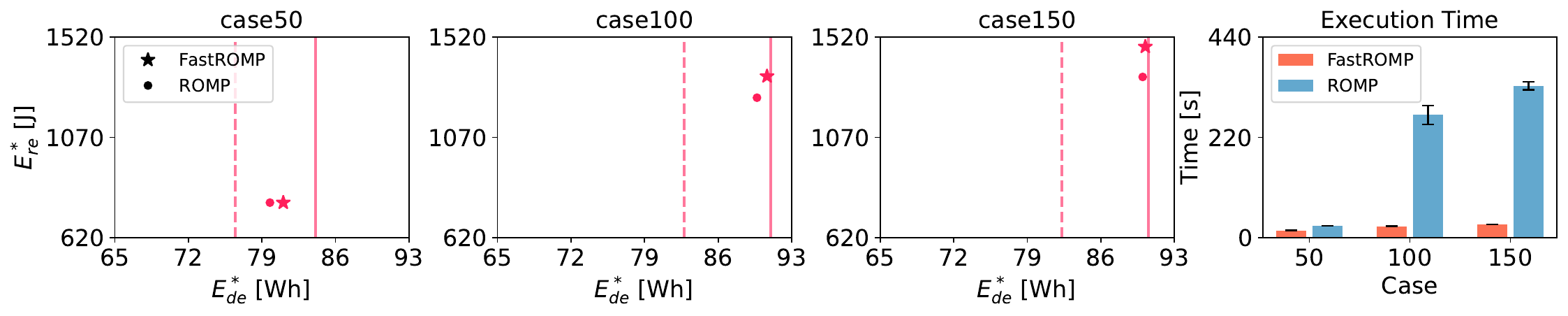}
        \caption{UAV energy budget is 110\% of the expected battery level.}
        \label{fig:dync-romp-bot}
    \end{subfigure}
    \caption{Performance comparison between FastROMP and ROMP in three cases, namely case 50, 100, and 150, where (a) and (b) refer to two scenarios in which the UAV charged ten sensor nodes, and its actual battery (i.e., energy budget, as presented by vertical solid lines) differs from the expected battery level (as presented by vertical dashed lines). The two sub-figures on the right represent the execution time required for our approaches under the above scenarios.}
    \label{fig:off-on-comp}
\end{figure*}

To better understand the behaviour of ROMP under dynamic updates, our initial focus is to analyze the number of iterations required for convergence toward the first feasible solution. Fig.~\ref{fig:sp-bgt} shows that the original step reduction loop takes 4222.46 s to converge towards the initial feasible solution. We employ a simple piece-wise function to accelerate this convergence process, effectively leveraging the difference between the estimated and desired discharged energy. Notably, we set a computational time limit of ten seconds for the GLS to balance the objectives of solution accuracy and computation time. Although this speedup strategy can significantly decrease the total number of iterations, the solution quality is not guaranteed. Multiple additional iterations may be needed for the best prize budget.

Accordingly, we propose FastROMP that eliminates the need to re-execute the `Initial Solver' by directly adjusting the previous route. To evaluate the proposed scheme, three distinct cases (case 50, 100, and 150 from \S~\ref{sec:iter}) were selected to represent the sparsely, moderately and densely distributed deployment of a WRSN, respectively. All experimental results were averaged over 20 executions. Fig.~\ref{fig:off-on-comp} illustrates the variations in the UAV's discharged and charged energy in response to two scenarios wherein the UAV charged ten sensor nodes, and its actual battery was 10\% lower (and higher) than the expected level. The expected battery levels of the three cases considered here were 76.4435 Wh, 82.7459 Wh, and 82.3198 Wh, respectively. Note that though case 100 and case 150 have different sensor densities, their expected battery levels are similar. This is because both cases are close to the upper bound of UAV discharged energy (for charging ten nodes in this grid size). Results indicate that ROMP can offer better solution quality than FastROMP when the UAV can charge all remaining sensor nodes. For instance, in case 50 of Fig.~\ref{fig:dync-romp-bot}, FastROMP's route has similar recharged energy as the ROMP's but more discharged energy (1.59\%) than ROMP's. While in other cases, FastROMP effectively utilized the energy budget to improve the solution quality. Notably, in case 100 of Fig.~\ref{fig:dync-romp-top}, FastROMP's route can obtain a gain of recharged energy (21.47\%) with a minor cost of UAV discharged energy (1.89\%), compared to ROMP's route. Moreover, as presented in Fig.~\ref{fig:off-on-comp}, FastROMP significantly reduces computation time by employing an online adjustment on the previous route. Depending on scenarios of whether the energy budget decreased (Fig.~\ref{fig:dync-romp-top}) or increased (Fig.~\ref{fig:dync-romp-top}), FastROMP demonstrated an increasing trend in saved execution time, ranging from 39.57\% to 93.3\%. This corresponds to a sensor node deployment from sparse to dense distribution.

\subsection{Complexity Analysis}
In the described tests, a two-dimensional vector with memory size $\big(N_{pop} \cdot N_{route}\big)$ is used for recording population solutions. As for the time complexity of CBHA, it takes $\big(N_{route} \cdot N_{sn}\big)$ iterations to complete the `attraction' phase, which is the most computationally demanding function in CBHA's evolution loop. The worst-case iterations taken for the main loop, therefore, should be $\big(N_{sol} \cdot N_{sn} \cdot N_{gen} \cdot N_{pop}\big)$. Note that $N_{sol} = N_{sn}$ in TSP scenarios, which takes $\big(N_{sn}^2 \cdot N_{gen} \cdot N_{pop}\big)$ iterations to compute. Therefore, CBHA has a time complexity of $O(n) \sim O(n^2)$ when $N_{sn}$ is considered the main input. For FastROMP, both operators require three vectors, each with a memory size $N_{route}$, to keep track of the UAV's route, energy, and flight time. During the computation of the add value list, the add operator requires a maximum of $\big( (N - N_{route})^2 \cdot N_{route} \big)$ iterations, which is the most computationally expensive part of the algorithm. The drop operator needs at most $N_{route}^2$ iterations to calculate the drop value list. As a result, the sequential execution of both operators will lead the FastROMP's worst-case time complexity to be $O(n^2)$.

\section{Conclusion and Future Work}
This paper presents a lightweight and reliable planner, ROMP, to solve the charging scheduling problem for a wireless rechargeable sensor network system. Characterized as a hybrid of dynamic TSP and OP scenarios, the optimization objective is to maximize the recharged energy of sensor nodes within the UAV energy constraint. We first develop an energy consumption model for M100 drones that considers flight regimes in the wind and IPT. The offline mission planning scheme involves an `Initial Solver' and an `Optimization Solver', which employs a Context-aware BHA to improve the initial solution from GLS. Implementing a straightforward parallelization strategy exploiting multiple processing cores further improves the performance of the offline scheme. Results show that using four processes, our distributed algorithm can achieve better solutions with a percentage of 99.8\% (over 500 executions) than GLS at 5000 populations. More importantly, it demonstrates that the system can converge in sufficient time for deployment in a realistic scenario, as shown on a typical flight computer. Results indicate that even in a TSP scenario, the high-quality solutions provided by GLS can be further improved by ROMP in windy environments. However, the step reduction loop in `Initial Solver' is computationally expensive and may not be applicable in the real-world scenario in time. Therefore, we developed FastROMP, which employs a new online adjustment operator that takes the latest state information as input. Exploiting the prior route from the last state, FastROMP is shown to deliver updated routes that better utilize the UAV's energy budget and charge more sensor nodes with a significant computational time reduction, compared to ROMP. 

Our future work will focus on improving the reliability and efficiency of ROMP and demonstrating the efficacy of the planner using UAV and sensor hardware at the appropriate scale. Field and wind tunnel experiments will be conducted to achieve robust performance and algorithm parameters. We will expand model parameters to include dynamic estimates of drag coefficient, fluid mass density, tilt angle changes, etc. We will also investigate the use of machine learning techniques, such as Long Short-Term Memory (LSTM), to predict energy consumption. By training a power consumption model with a data-driven LSTM neural network, we may reduce the computational expense for the onboard planner~\cite{SHE2020LSTMPOWER}. Moreover, we will explore the simulation of various event-based sensor types in WRSNs, including motion and audio sensors. These sensors typically exhibit a stochastic energy depletion rate, which is contingent upon the distribution of detected events. This characteristic can result in a stochastic `prize' associated with visiting a node. Therefore, the scheduling problem presents a compelling opportunity for robust optimization.

\ifCLASSOPTIONcaptionsoff
  \newpage
\fi

\bibliographystyle{ieeetr}
\bibliography{ref.bib}

\end{document}